\documentclass[final,10pt]{IEEEtran}
\usepackage{times}
\usepackage{epsfig}
\usepackage{graphicx}
\usepackage{amssymb}
\usepackage{amsmath}
\usepackage{enumerate}
\usepackage{indentfirst}
\usepackage{algorithmic}
\usepackage{algorithm}
\usepackage{url}
\usepackage{booktabs}
\usepackage{multirow}
\usepackage{longtable}
\usepackage{cite}
\usepackage{amsmath}
\usepackage{tkz-graph}
\usepackage{graphicx}
\usetikzlibrary{calc}
\usepackage{tkz-euclide}
\usepackage{caption}
\usepackage{subcaption}
\usepackage{tikz}
\usepackage{hyperref}
\usepackage{epstopdf}
\usepackage{multicol}
\usepackage{tabularx}
\usepackage{nccmath}
\usepackage{stfloats}
\usepackage{subcaption}

\newcommand{\xbf}{\boldsymbol{x}}

\newcommand{\Gammabf}{\boldsymbol{\Gamma}}

\newcommand{\HRule}[1][\medskipamount]{\par
  \vspace*{\dimexpr-\parskip-\baselineskip+#1}
  \noindent\rule{\linewidth}{0.2mm}\par
  \vspace*{\dimexpr-\parskip-.5\baselineskip+#1}}

\newcommand{\argmin}{\arg\!\min}
\newcommand{\argmax}{\arg\!\max}

\long\def\symbolfootnote[#1]#2{\begingroup\def\thefootnote{\fnsymbol{footnote}}
\footnote[#1]{#2}\endgroup}
\ifCLASSINFOpdf
\else
\fi
\begin{document}
%
\title{A GAMP Based Low Complexity Sparse Bayesian Learning Algorithm}

\author{Maher~Al-Shoukairi,~\IEEEmembership{Student Member,~IEEE,}
        Philip~Schniter,~\IEEEmembership{Fellow,~IEEE,}
        and~Bhaskar~D.~Rao,~\IEEEmembership{Fellow,~IEEE}
\thanks{M. Al-Shoukairi and B. Rao (E-mail: malshouk,brao@ucsd.edu) are with the Department of Electrical and Computer Engineering, University of California, San Diego, La Jolla, California. Their work is supported by the Ericsson Endowed Chair funds}%
\thanks{P. Schniter (E-mail: schniter@ece.osu.edu) is with the Department of Electrical
and Computer Engineering, The Ohio State University, Columbus, Ohio. His work is supported by the National Science Foundation grant 1527162}
}

\maketitle

\begin{abstract}
In this paper, we present an algorithm for the sparse signal recovery problem that incorporates damped Gaussian generalized approximate message passing (GGAMP) into Expectation-Maximization (EM)-based sparse Bayesian learning (SBL). In particular, GGAMP is used to implement the E-step in SBL in place of matrix inversion, leveraging the fact that GGAMP is guaranteed to converge with appropriate damping. The resulting GGAMP-SBL algorithm is much more robust to arbitrary measurement matrix $\boldsymbol{A}$ than the standard damped GAMP algorithm while being much lower complexity than the standard SBL algorithm. We then extend the approach from the single measurement vector (SMV) case to the temporally correlated multiple measurement vector (MMV) case, leading to the GGAMP-TSBL algorithm. We verify the robustness and computational advantages of the proposed algorithms through numerical experiments.\end{abstract}
\IEEEpeerreviewmaketitle

\section{Introduction}
\subsection{Sparse Signal Recovery}
The problem of sparse signal recovery (SSR) and the related problem of compressed sensing have received much attention in recent years \cite{donoho2006compressed,candes2006near,baraniuk2007compressive,Elad201008,FoucartRauhut201306,EldarKutyniok201206}.
The SSR problem,
in the single measurement vector (SMV) case, consists of recovering a sparse signal $\boldsymbol{x} \in \mathbb{R}^N$ from $M \leq N$ noisy linear measurements $\boldsymbol{y} \in \mathbb{R}^M$:
\begin{flalign}
&\boldsymbol{y=Ax+e,} \label{eq:SMVmodel}
\end{flalign}
where $\boldsymbol{A} \in \mathbb{R}^{M \times N}$ is a known measurement matrix and $\boldsymbol{e} \in \mathbb{R}^M$ is additive noise modeled by $ \boldsymbol{e} \sim \mathcal{N} (0,\sigma^2 \boldsymbol{I})$.
Despite the difficulty in solving this problem \cite{natarajan1995sparse}, an important finding in recent years is that for a sufficiently sparse $\boldsymbol{x}$ and a well designed $\boldsymbol{A}$, accurate recovery is possible by techniques  such as basis pursuit and orthogonal matching pursuit \cite{candes2005decoding,chen2001atomic,tropp2007signal}. The SSR problem has seen considerable advances on the algorithmic front and they include iteratively reweighted algorithms \cite{figueiredo2007majorization,candes2008enhancing,chartrand2008iteratively} and Bayesian techniques \cite{tipping2001sparse,wipf2004sparse,vila2013expectation,babacan2010bayesian,he2009exploiting,ji2008bayesian,baron2010bayesian}, among others. Two Bayesian techniques related to this work are the generalized approximate message passing (GAMP) and the sparse Bayesian learning (SBL) algorithms. We briefly discuss both algorithms and some of their shortcomings that we intend to address in this work.

\subsection{Generalized Approximate Message Passing Algorithm}
Approximate message passing (AMP) algorithms apply quadratic and Taylor series approximations to loopy belief propagation to produce low complexity algorithms. Based on the original AMP work in \cite{donoho2010message}, a generalized AMP (GAMP) algorithm was proposed in \cite{rangan2011generalized}. The GAMP algorithm provides an iterative Bayesian framework under which the knowledge of the matrix $\boldsymbol{A}$ and the densities $p(\boldsymbol{x})$ and $p(\boldsymbol{y} \mathcal{\vert} \boldsymbol{x})$ can be used to compute the maximum a posteriori (MAP) estimate $\hat{\boldsymbol{x}}_{MAP}= {\arg\min}_{\boldsymbol{x} \in \mathbb{R}^{N}} p(\boldsymbol{x} \mathcal{\vert} \boldsymbol{y})$ when it is used in its max-sum mode, or approximate the minimum mean-squared error (MMSE) estimate $\hat{\boldsymbol{x}}_{MMSE}=\int_{\mathbb{R}^{N}} \boldsymbol{x} p(\boldsymbol{x} \mathcal{\vert} \boldsymbol{y}) d\boldsymbol{x} = \mathbb{E}(\boldsymbol{x} \mathcal{\vert} \boldsymbol{y})$ when it is used in its sum-product mode.

The performance of AMP/GAMP algorithms in the large system limit ($M,N \rightarrow \infty$) under an i.i.d zero-mean sub-Gaussian matrix $\boldsymbol{A}$ is characterized by state evolution \cite{bayati2011dynamics}, whose fixed points, when unique, coincide with the MAP or the MMSE estimate. However, when $\boldsymbol{A}$ is generic, GAMP's fixed points can be strongly suboptimal and/or the algorithm may never reach its fixed points. Previous work has shown that even mild ill-conditioning or small mean perturbations in $\boldsymbol{A}$ can cause GAMP to diverge \cite{Schniter_conv,caltagirone2014convergence,MADGAMP}. To overcome the convergence problem in AMP algorithms, a number of AMP modifications have been proposed. A ``swept`` GAMP (SwAMP) algorithm was proposed in \cite{swept}, which replaces parallel variable updates in the GAMP algorithm with serial ones to enhance convergence. But SwAMP is relatively slow and still diverges for certain $\boldsymbol{A}$. An adaptive damping and mean-removal procedure for GAMP was proposed in \cite{MADGAMP} but it too is somewhat slow and still diverges for certain $\boldsymbol{A}$. An alternating direction method of multipliers (ADMM) version of AMP was proposed in \cite{rangan2015inference} with improved robustness but even slower convergence.

In the special case that the prior and likelihood are both independent Gaussian, \cite{Schniter_conv} was able to provide a full characterization of GAMP's convergence. In particular, it was shown that Gaussian GAMP (GGAMP) algorithm will converge if and only if the peak to average ratio of the squared singular values in $\boldsymbol{A}$ is sufficiently small. When this condition is not met, \cite{Schniter_conv} proposed a damping technique that guarantees convergence of GGAMP at the expense of slowing its convergence rate. Although the case of Gaussian prior and likelihood is not enough to handle the sparse signal recovery problem directly, it is sufficient to replace the costly matrix inversion step of the standard EM-based implementation of SBL, as we describe below.

\subsection{Sparse Bayesian Learning Algorithm}
To understand the contribution of this paper, we give a very brief description of SBL \cite{tipping2001sparse,wipf2004sparse}, saving a more detailed description for Section II. Essentially, SBL is based on a Gaussian scale mixture (GSM) \cite{andrews1974scale,palmer2005variational,lange1993normal} prior on $\boldsymbol{x}$. That is, the prior is Gaussian conditioned on a variance vector $\boldsymbol{\gamma}$, which  is then controlled by a suitable choice of hyperprior $p(\boldsymbol{\gamma}).$
A large of  number of sparsity-promoting priors, like the Student-t and Laplacian priors, can be modeled using a GSM, making the approach widely applicable \cite{andrews1974scale,lange1993normal,palmer2005variational,pedersen2015sparse}. In the SBL algorithm, the expectation-maximization (EM)  algorithm is used to alternate between estimating $\boldsymbol{\gamma}$ and estimating the signal $\boldsymbol{x}$ under fixed $\boldsymbol{\gamma}$. Since the latter step uses a Gaussian likelihood and Gaussian prior, the exact solution can be computed in closed form via matrix inversion. This matrix inversion is computationally expensive, limiting the algorithms applicability to large scale problems.

\subsection{Paper's Contribution \protect\footnote{Part of this work was presented in the 2014 Asilomar conference on Signals, Systems and Computers \cite{AMPSBL}}}

In this paper, we develop low-complexity algorithms for sparse Bayesian learning (SBL) \cite{tipping2001sparse,wipf2004sparse}.
Since the traditional implementation of SBL uses matrix inversions at each iteration, its complexity is too high for large-scale problems. In this paper we circumvent the matrix inverse using the generalized approximate message passing (GAMP) algorithm \cite{rangan2011generalized,donoho2010message,Schniter_conv}. Using GAMP to implement the E step of EM-based SBL provides a significant reduction in complexity over the classical SBL algorithm. This work is a beneficiary of the algorithmic improvements and theoretical insights that have taken place in recent work in AMP \cite{donoho2010message,rangan2011generalized,Schniter_conv}, where we exploit the fact that using a Gaussian prior on $p(\boldsymbol{x})$ can provide guarantees for the GAMP E-step to not diverge when sufficient damping is used \cite{Schniter_conv}, even for a non-i.i.d.-Gaussian $\boldsymbol{A}$. In other words, the enhanced robustness of the proposed algorithm is due to the GSM prior used on $\boldsymbol{x}$, as opposed to other sparsity promoting priors for which there are no GAMP convergence guarantees when $\boldsymbol{A}$ is non-i.i.d.-Gaussian. The resulting algorithm is the Gaussian GAMP SBL (GGAMP-SBL) algorithm, which combines the robustness of SBL with the speed of GAMP.


To further illustrate and expose the synergy between the AMP and SBL frameworks, we also propose a new approach to the multiple measurement vector (MMV) problem. The MMV problem extends the SMV problem from a single measurement and signal vector to a sequence of measurement and signal vectors. Applications of MMV include direction of arrival (DOA) estimation and EEG/MEG source localization, among others. In our treatment of the MMV problem, all signal vectors are assumed to share the same support. In practice it is often the case that the non-zero signal elements will experience temporal correlation, i.e., each non-zero row of the signal matrix can be treated as a correlated time series. If this correlation is not taken into consideration, the performance of MMV algorithms can degrade quickly \cite{zhang2011sparse}. Extensions of SBL to the MMV problem have been developed in \cite{wipf2007empirical,zhang2011sparse,prasad2014joint}, such as the TSBL and TMSBL algorithms \cite{zhang2011sparse}. Although TMSBL has lower complexity than TSBL, it still requires an order of $\mathcal{O} (N M^2)$ operations per iteration, making it unsuitable for large-scale problems. To overcome the complexity problem, \cite{AMP_MMV} and \cite{AMPSBL} proposed AMP-based Bayesian approaches to the MMV problem. However, similar to the SMV case, these algorithms are only expected to work for i.i.d zero-mean sub-Gaussian $\boldsymbol{A}$. We therefore extend the proposed GGAMP-SBL to the MMV case, to produce a GGAMP-TSBL algorithm that is more robust to generic $\boldsymbol{A}$, while achieving linear complexity in all problem dimensions.

The organization of the paper is as follows. In Section~\ref{Emp_Bayes}, we review SBL. In Section \ref{DGAMP_SBL}, we combine the damped GGAMP algorithm with the SBL approach to solve the SMV problem and investigate its  convergence behavior. In Section \ref{DGAMP_TSBL}, we use a time-correlated multiple measurement factor graph to derive the GGAMP-TSBL algorithm.
In Section \ref{Numerical_Results}, we present numerical results to compare the performance and complexity of the proposed algorithms with the original SBL and with other AMP algorithms for the SMV case, and with TMSBL for the MMV case. The results show that the new algorithms maintained the robustness of the original SBL and TMSBL algorithms, while significantly reducing their complexity.



\section{Sparse Bayesian Learning for SSR}\label{Emp_Bayes}
\label{EMp_Bayes}

\subsection{GSM Class of Priors}
We will assume that the entries of $\xbf$ are independent and identically distributed, i.e. $p(\boldsymbol{x}) = \Pi_n p(x_n).$
The sparsity promoting prior $p(x_n)$ will be chosen from the GSM class and so will admit the following representation
\begin{flalign}
\label{eq:prior}
&p(x_n) = \int{\mathcal{N}(x_n;0,\gamma_n) p({\gamma_n}) d {\gamma_n}},
\end{flalign}
where $\mathcal{N}(x_n;0,\gamma_n)$  denotes a Gaussian density with mean zero and variance $\gamma_n$. The mixing density on hyperprior $p({\gamma_n})$ controls the prior on $x_n$. For instance,
if a Laplacian prior is desired for $x_n$, then an exponential density is chosen for $p(\gamma_n)$ \cite{andrews1974scale}.

In the empirical Bayesian approach,
an estimate of the hyperparameter vector $\boldsymbol{\gamma}$ is iteratively estimated, often using evidence maximization.
For a given estimate $\hat{\boldsymbol{\gamma}}$, the posterior $p(\boldsymbol{x} \mathcal{\vert} \boldsymbol{y})$ is approximated as $p(\boldsymbol{x} \mathcal{\vert} \boldsymbol{y} ; \hat{\boldsymbol{\gamma}})$, and the mean of this posterior is used as a point estimate for $\hat{\boldsymbol{x}}$.
This mean can be computed in closed form, as detailed below, because the approximate posterior is Gaussian.
It was shown in \cite{ritwik} that this empirical Bayesian method, also referred to as a Type II maximum likelihood method, can be used to formulate a number of algorithms for solving the SSR problem by changing the mixing density $p(\gamma_n)$.
There are many computational methods that can be employed for computing $\boldsymbol{\gamma}$ in the evidence maximization framework, e.g, \cite{tipping2001sparse,tipping2003fast,wipf2004sparse}. In this work, we utilize the EM-SBL algorithm because of its synergy with the GAMP framework, as will be apparent below.
\subsection{SBL's EM Algorithm}
In EM-SBL, the EM algorithm is used to learn the  unknown signal variance vector $\boldsymbol{\gamma}$ \cite{Bishop2006pattern,mclachlan2007algorithm,wu1983convergence}, and possibly also the noise variance $\sigma^2$.
We focus on learning $\boldsymbol{\gamma}$, assuming the noise variance $\sigma^2$ is known. We later state the EM update rule for the noise variance $\sigma^2$ for completeness.

The goal of the EM algorithm is to maximize the posterior  $p(\boldsymbol{\gamma} \mathcal{\vert} \boldsymbol{y})$ or equivalently\footnote{Using Bayes rule, $p(\boldsymbol{\gamma} \mathcal{\vert} \boldsymbol{y})$ = $p(\boldsymbol{y}, \boldsymbol{\gamma}) / p(\boldsymbol{y})$ where $p(\boldsymbol{y})$ is a constant with respect to $\boldsymbol{\gamma}$. Thus for MAP estimation of $\boldsymbol{\gamma}$ we can maximize $p(\boldsymbol{y}, \boldsymbol{\gamma})$, or minimize $- \log p(\boldsymbol{y}, \boldsymbol{\gamma})$.}
to minimize $-\log p(\boldsymbol{y},\boldsymbol{\gamma})$.
For the GSM prior \eqref{eq:prior} and the additive white Gaussian noise model,  this results in the SBL cost function \cite{tipping2001sparse,wipf2004sparse},
\begin{flalign}
\label{eq:cost} \chi(\boldsymbol{\gamma})&= -\log p(\boldsymbol{y}, \boldsymbol{\gamma}) \nonumber \\ &= \frac{1}{2} \log |\boldsymbol{\Sigma_y}| + \frac{1}{2}  \boldsymbol{y}^\top \boldsymbol{\Sigma^{-1}_y} \boldsymbol{y} - \log p(\boldsymbol{\gamma}),\\
\boldsymbol{\Sigma_y}&= \sigma^2 \boldsymbol{I} +\boldsymbol{A} \boldsymbol{\Gamma} \boldsymbol{A}^\top, \ \ \Gammabf \triangleq \text{Diag}(\boldsymbol{\gamma}). \nonumber
\end{flalign}

In the EM-SBL approach,  $\boldsymbol{x}$ is treated as the hidden variable and the parameter estimate is iteratively updated as follows:
 \begin{flalign}
\boldsymbol{{\gamma}}^{i+1} = \argmax_{\boldsymbol{\gamma}} \mathbb{E}_{\boldsymbol{x} \mathcal{\vert} \boldsymbol{y} ; \boldsymbol{\gamma}^{i}} \left[ \log p(\boldsymbol{y}, \boldsymbol{x} , \boldsymbol{\gamma})\right], \label{eq:gamma_iteration}
\end{flalign}
where $p(\boldsymbol{y}, \boldsymbol{x} , \boldsymbol{\gamma})$ is the joint probability of the complete data and $p({\boldsymbol{x} \mathcal{\vert} \boldsymbol{y} ; \boldsymbol{\gamma}^{i}})$ is the  posterior under the old parameter estimate $\boldsymbol{\gamma}^{i},$ which is used to evaluate the expectation. In each iteration, an expectation has to be computed (E-step) followed by a maximization step (M-step).
It is easy to show that at each iteration, the EM algorithm increases a lower bound on the log posterior $\log p(\boldsymbol{\gamma} \mathcal{\vert} \boldsymbol{y})$ \cite{Bishop2006pattern}, and it has been shown in \cite{wu1983convergence} that the algorithm will converge to a stationary point
of the posterior under a fairly general set of conditions that are applicable in many practical applications.

Next we detail the implementation of the E and M steps of the EM-SBL algorithm.
\subsubsection*{\textbf{SBL's E-step}}
The Gaussian assumption on the additive noise $\boldsymbol{e}$ leads to the following Gaussian likelihood function:
\begin{flalign}
&p(\boldsymbol{y} \mathcal{\vert} \boldsymbol{x};\sigma^2)= \frac{1}{(2 \pi \sigma^2)^{\frac{M}{2}}} \exp \left(-\frac{1}{2 \sigma^2} \| \boldsymbol{y-Ax} \|^2 \right). \label{eq:py}
\end{flalign}
Due to the GSM prior \eqref{eq:prior}, the density of $\boldsymbol{x}$ conditioned on $\boldsymbol{\gamma}$ is Gaussian:
\begin{flalign}
&p(\boldsymbol{x} \mathcal{\vert} \boldsymbol{\gamma})=\prod_{n=1}^{N}\frac{1}{(2 \pi \gamma_n)^{\frac{1}{2}}} \exp \left(-\frac{x_n^2}{2 \gamma_n} \right). \label{eq:px}
\end{flalign}
Putting \eqref{eq:py} and \eqref{eq:px} together, the density needed for the E-step is Gaussian:
\begin{flalign}
p(\boldsymbol{x} \mathcal{\vert} \boldsymbol{y} ,\boldsymbol{\gamma})&= \mathcal{N}(\boldsymbol{x}; \hat{\boldsymbol{x}}, \boldsymbol{\Sigma_x}) \label{eq:posterior}\\
\hat{\boldsymbol{x}}&=\sigma^{-2} \boldsymbol{\Sigma_x} \boldsymbol{A}^\top \boldsymbol{y} \label{eq:mux}\\
\boldsymbol{\Sigma_x}&=(\sigma^{-2} \boldsymbol{A}^\top \boldsymbol{A} + \boldsymbol{\Gamma}^{-1})^{-1} \nonumber \\ &=\boldsymbol{\Gamma}-\boldsymbol{\Gamma} \boldsymbol{A}^\top (\sigma^2  \boldsymbol{I}+\boldsymbol{A} \boldsymbol{\Gamma} \boldsymbol{A}^\top)^{-1} \boldsymbol{A} \boldsymbol{\Gamma}. \label{eq:Sigmax}
\end{flalign}
We refer to the mean vector as $\hat{\boldsymbol{x}}$ since it will be used as the SBL point estimate of $\boldsymbol{x}$. In the sequel, we will use $\boldsymbol{\tau}_x$ when referring to the vector composed from the diagonal entries of the covariance matrix $\boldsymbol{\Sigma_x}$. Although both $\hat{\boldsymbol{x}}$ and $\boldsymbol{\tau}_x$ change with the iteration $i$, we will sometimes omit their $i$ dependence for brevity. Note that the mean and covariance computations in \eqref{eq:mux} and \eqref{eq:Sigmax} are not affected by the choice of $p(\boldsymbol{\gamma})$. The mean and diagonal entries of the covariance matrix are needed to carry out the M-step as shown next.
\subsubsection*{\textbf{SBL's M-Step}}
The M-step is then carried out as follows. First notice that
\small
\begin{flalign}
&\mathbb{E}_{\boldsymbol{x} \mathcal{\vert} \boldsymbol{y} ; \boldsymbol{\gamma}^{i}, \sigma^2} \left[-\log p(\boldsymbol{y}, \boldsymbol{x} , \boldsymbol{\gamma}; \sigma^2)\right]=
\nonumber \\ &\mathbb{E}_{\boldsymbol{x} \mathcal{\vert} \boldsymbol{y} ; \boldsymbol{\gamma}^{i}, \sigma^2} \left[-\log p(\boldsymbol{y} \mathcal{\vert} \boldsymbol{x} ; \sigma^2) - \log p(\boldsymbol{x} \mathcal{\vert} \boldsymbol{\gamma}) -\log p(\boldsymbol{\gamma})\right]. \label{eq:Estep1}
\end{flalign}
\normalsize
Since the first term in \eqref{eq:Estep1} does not depend on $\boldsymbol{\gamma}$, it will not be relevant for the M-step and thus can be ignored. Similarly, in the subsequent steps we will drop constants and terms that do not depend on $\boldsymbol{\gamma}$ and therefore do not impact the M-step. Since $ \mathbb{E}_{\boldsymbol{x} \mathcal{\vert} \boldsymbol{y} ; \boldsymbol{\gamma}^{i}, \sigma^2}[x_n^2]= \hat{x}_n^2 + \tau_{x_n}$,
\begin{flalign}
&\mathbb{E}_{\boldsymbol{x} \mathcal{\vert} \boldsymbol{y} ; \boldsymbol{\gamma}^{i}, \sigma^2}  [ -\log p(\boldsymbol{x} \mathcal{\vert} \boldsymbol{\gamma}) -\log p(\boldsymbol{\gamma}) ] = \nonumber\\ &\sum_{n=1}^N \left ( \left (\frac{ \hat{x}_n^2 +\tau_{x_n}}{2\gamma_n}\right) + \frac{1}{2} \log \gamma_n - \log p({\gamma_n})\right). \label{eq:Estep2}
\end{flalign}
Note that the E-step only requires $\hat{x}_n,$ the posterior mean from \eqref{eq:mux}, and $\tau_{x_n}, $the posterior variance from \eqref{eq:Sigmax}, which are statistics of the marginal densities $p(x_n|\boldsymbol{y},\boldsymbol{\gamma}^i)$. In other words, the full joint posterior $p(\boldsymbol{x} \mathcal{\vert} \boldsymbol{y},\boldsymbol{\gamma}^i)$ is not needed. This facilitates the use of message passing algorithms.

As can be seen from \eqref{eq:posterior}-\eqref{eq:Sigmax}, the computation of $\hat{\boldsymbol{x}}$ and $\boldsymbol{\tau}_x$ involves the inversion of an $N \times N$ matrix, which can be reduced to $M \times M$ matrix inversion by the matrix inversion lemma. The complexity of computing $\hat{\boldsymbol{x}}$ and $\boldsymbol{\tau}_x$ can be shown to be $ \mathcal{O} (N M^2)$ under the assumption that $M \leq N$. This makes the EM-SBL algorithm computationally prohibitive and impractical to use with large dimensions.

From \eqref{eq:gamma_iteration} and \eqref{eq:Estep2}, the M-step for each iteration is as follows:
\small
\begin{subequations}
\begin{flalign}
\boldsymbol{\gamma}^{i+1} &= \argmin_{\boldsymbol{\gamma}} \left [\sum_{n=1}^N \left (\frac{ \hat{x}_n^2 +\tau_{x_n}}{2\gamma_n} + \frac{\log \gamma_n}{2}  - \log p({\gamma_n})\right) \right ]. \label{eq:mstep}
\intertext{\normalsize This reduces to $N$ scalar optimization problems,}
\small
\gamma_n^{i+1}&=\argmin_{\gamma_n} \left [\frac{ \hat{x}_n^2 + \tau_{x_n}}{2\gamma_n} + \frac{1}{2} \log \gamma_n - \log p({\gamma_n}) \right ]. \label{eq:mstep2}
\end{flalign}
\end{subequations}
\normalsize
The choice of hyperprior $p(\boldsymbol{\gamma})$ plays a role in the M-step, and governs the prior for $\boldsymbol{x}$. However, from the computational simplicity of the M-step, as evident from (\ref{eq:mstep2}), the hyperprior rarely impacts the overall algorithmic computational complexity, which is mainly that of computing the quantities $\hat{\boldsymbol{x}}$ and $\boldsymbol{\tau}_x$ in the E-step.

Often a non-informative prior is used in SBL. For the purpose of obtaining the M-step update, we will also simplify and drop $p(\boldsymbol{\gamma})$ and  compute the Maximum Likelihood estimate of $\boldsymbol{\gamma}.$  From \eqref{eq:mstep2}, this reduces to,
$\gamma_n^{i+1} = \hat{x}_n^2 + \tau_{x_n}$.

Similarly, if the noise variance $\sigma^2$ is unknown, it can be estimated using:
\begin{flalign}
{{(\sigma^{2})}^{i+1}}&= \argmax_{\sigma^2} \mathbb{E}_{\boldsymbol{x} \mathcal{\vert} \boldsymbol{y} , \boldsymbol{\gamma}; {{(\sigma}^{2})}^i} [p(\boldsymbol{y}, \boldsymbol{x} , \boldsymbol{\gamma}; \sigma^2)] \nonumber\\
&=\frac{\| \boldsymbol{y} - \boldsymbol{A} \boldsymbol{x} \|^2 + {(\sigma^2)}^i \sum_{n=1}^N \left( 1- \frac{\tau_{x_n}}{\gamma_n} \right)}{M}. \label{eq:sigma2}
\end{flalign}
We note here that estimates obtained by \eqref{eq:sigma2} can be highly inaccurate as mentioned in \cite{wipf2007empirical}. Therefore, it suggests that experimenting with different values of $\sigma^2$ or using some other application based heuristic will probably lead to better results.

\section{Damped Gaussian GAMP SBL} \label{DGAMP_SBL}
We now show how damped GGAMP can be used to simplify the E-step above, leading to the damped GGAMP-SBL algorithm.
Then we examine the convergence behavior of the resulting algorithm.

\subsection{GGAMP-SBL}
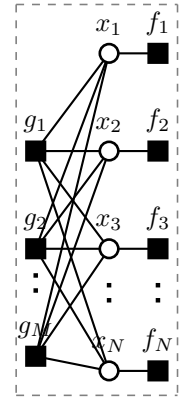
\begin{figure}[h]
\hspace*{-2em}\begin{tikzpicture}[scale=.65]
\useasboundingbox (0,-5) rectangle (-10,2.5);
\SetVertexNormal[MinSize = 7pt,Shape = rectangle, FillColor  = black, LineWidth = 1pt]
\foreach \a/\b in {.5/T}
{\Vertex[L=$g_1$,x=0+3*\a-5,y=0-\a,LabelOut,Lpos=90]{G1\b}
\Vertex[L=$g_2$,x=0+3*\a-5,y=-2-\a,LabelOut,Lpos=90]{G2\b}
\Vertex[L=$g_M$,x=0+3*\a-5,y=-4.2-\a,LabelOut,Lpos=90]{G3\b}

\draw[loosely dotted,line width=1.7pt] (3*\a-5,-2.5-\a) -- (3*\a-5,-3-\a);

\SetVertexNormal[MinSize = 7pt,LineWidth = 1pt]
\Vertex[L=$x_1$,x=1.5+3*\a-5,y=2-\a,LabelOut,Lpos=90]{X1\b}
\Vertex[L=$x_2$,x=1.5+3*\a-5,y=0-\a,LabelOut,Lpos=90]{X2\b}
\Vertex[L=$x_3$,x=1.5+3*\a-5,y=-2-\a,LabelOut,Lpos=90]{X3\b}
\Vertex[L=$x_N$,x=1.5+3*\a-5,y=-4.5-\a,LabelOut,Lpos=90]{X4\b}

\draw[loosely dotted,line width=1.7pt] (1.5+3*\a-5,-2.7-\a) -- (1.5+3*\a-5,-3.2-\a);

\SetVertexNormal[MinSize = 7pt,Shape = rectangle, FillColor  = black, LineWidth = 1pt]
\Vertex[L=$f_1$,x=2.5+3*\a-5,y=2-\a,LabelOut,Lpos=90]{F1\b}
\Vertex[L=$f_2$,x=2.5+3*\a-5,y=0-\a,LabelOut,Lpos=90]{F2\b}
\Vertex[L=$f_3$,x=2.5+3*\a-5,y=-2-\a,LabelOut,Lpos=90]{F3\b}
\Vertex[L=$f_N$,x=2.5+3*\a-5,y=-4.5-\a,LabelOut,Lpos=90]{F4\b}
\draw[loosely dotted,line width=1.7pt] (2.5+3*\a-5,-2.7-\a) -- (2.5+3*\a-5,-3.2-\a);

\Edge(G1\b)(X1\b)
\Edge(G2\b)(X1\b)
\Edge(G3\b)(X1\b)
\Edge(G1\b)(X2\b)
\Edge(G2\b)(X2\b)
\Edge(G3\b)(X2\b)
\Edge(G1\b)(X3\b)
\Edge(G2\b)(X3\b)
\Edge(G3\b)(X3\b)
\Edge(G1\b)(X4\b)
\Edge(G2\b)(X4\b)
\Edge(G3\b)(X4\b)
\Edge(X1\b)(F1\b)
\Edge(X2\b)(F2\b)
\Edge(X3\b)(F3\b)
\Edge(X4\b)(F4\b)

\tkzDefPoint(0+3*\a-5-.4,3-\a){A\b}
\tkzDefPoint(0+3*\a-5-.4,-5-\a){B\b}
\tkzDefPoint(2.7+3*\a-5+.2,3-\a){C\b}
\tkzDefPoint(2.7+3*\a-5+.2,-5-\a){D\b}
\tkzDrawSegments[style=dashed,color=gray](A\b,B\b)
\tkzDrawSegments[style=dashed,color=gray](C\b,D\b)
\tkzDrawSegments[style=dashed,color=gray](A\b,C\b)
\tkzDrawSegments[style=dashed,color=gray](B\b,D\b)}

\end{tikzpicture}
\\
\caption{GAMP Factor Graph}
\label{SMV_factor_graph}
\end{figure}

Above we showed that, in the EM-SBL algorithm, the M-step is computationally simple but the E-step is computationally demanding. The GAMP algorithm can be used to efficiently approximate the quantities $\hat{\boldsymbol{x}}$ and $\boldsymbol{\tau}_x$ needed in the E-step, while the M-step remains unchanged. GAMP is based on the factor graph in Figure~\ref{SMV_factor_graph}, where for a given prior $f_n(x)=p(x_n)$ and a likelihood function $g_m=p(y_m \mathcal{\vert} \boldsymbol{x})$, GAMP uses quadratic approximations and Taylor series expansions, to provide approximations of MAP or MMSE estimates of $\boldsymbol{x}$. The reader can refer to \cite{rangan2011generalized} for detailed derivation of GAMP.
The E-step in Table~\ref{tab:SMV_algorithm}, uses the damped GGAMP algorithm from \cite{Schniter_conv} because of its ability to enhance traditional GAMP algorithm divergence issues with non-i.i.d.-Gaussian $\boldsymbol{A}$. The damped GGAMP algorithm has an important modification over the original GAMP algorithm and also over the previously proposed AMP-SBL \cite{AMPSBL}, namely the introduction of damping factors $\theta_s, \theta_x \in (0,1]$ to slow down updates and enhance convergence. Setting $\theta_s = \theta_x = 1$ in the damped GGAMP algorithm will yield no damping, and reduces the algorithm to the original GAMP algorithm. We note here that the damped GGAMP algorithm from \cite{Schniter_conv} is referred to by GGAMP, and therefore we will be using the terms GGAMP and damped GGAMP interchangeably in this paper. Moreover, when the components of the matrix $\boldsymbol{A}$ are not zero-mean, one can incorporate the same mean removal technique used in \cite{MADGAMP}.
The input and output functions $g_s(\boldsymbol{p},\boldsymbol{\tau}_p)$ and $g_x(\boldsymbol{r},\boldsymbol{\tau}_r)$ in Table~\ref{tab:SMV_algorithm} are defined based on whether the max-sum or the sum-product version of GAMP is being used. The intermediate variables $\boldsymbol{r}$ and $\boldsymbol{p}$ are interpreted as approximations of Gaussian noise corrupted versions of $\boldsymbol{x}$ and $\boldsymbol{z}=\boldsymbol{Ax}$, with the respective noise levels of $\boldsymbol{\tau}_r$ and $\boldsymbol{\tau}_p$.
In the max-sum version, the vector MAP estimation problem is reduced to a sequence of scalar MAP estimates given $\boldsymbol{r}$ and $\boldsymbol{p}$ using the input and output functions, where they are defined as:
\begin{flalign}
&[g_s(\boldsymbol{p},\boldsymbol{\tau}_p)]_m= p_m - \tau_{p_m} \text{prox}_{ \frac{-1}{\tau_{p_m}} \ln p(y_m \mathcal{\vert} z_m)} (\frac{p_m}{\tau_{p_m}}) \label{gx_max_sum}\\
&[g_x(\boldsymbol{r},\boldsymbol{\tau}_r)]_n= \text{prox}_{- \tau_{r_n} \ln p(x_n)} (r_n) \label{gs_max_sum}\\
&\text{prox}_{f} (r) \triangleq \argmin_{x} f(x)+ \frac{1}{2} |x-r |^2.
\end{flalign}

Similarly, in the sum-product version of the algorithm, the vector MMSE estimation problem is reduced to a sequence of scalar MMSE estimates given $\boldsymbol{r}$ and $\boldsymbol{p}$ using the input and output functions, where they are defined as:
\begin{flalign}
&[g_s(\boldsymbol{p},\boldsymbol{\tau}_p)]_m= \frac{\int z_m p(y_m \mathcal{\vert} z_m) \mathcal{N}(z_m;\frac{p_m}{\tau_{p_m}},\frac{1}{\tau_{p_m}}) dz_m }{\int p(y_m \mathcal{\vert} z_m) \mathcal{N}(z_m;\frac{p_m}{\tau_{p_m}},\frac{1}{\tau_{p_m}}) dz_m}  \label{gs_sum_product} \\
&[g_x(\boldsymbol{r},\boldsymbol{\tau}_r)]_n= \frac{\int x_n p(x_n) \mathcal{N}(x_n;r_n,\tau_{r_n}) dx_n }{\int p(x_n) \mathcal{N}(x_n;r_n,\tau_{r_n}) dx_n}. \label{gx_sum_product}
\end{flalign}

For the parametrized Gaussian prior we imposed on $\boldsymbol{x}$ in \eqref{eq:prior}, both sum-product and max-sum versions of $g_x(\boldsymbol{r},\boldsymbol{\tau}_r)$ yield the same updates for $\hat{\boldsymbol{x}}$ and $\boldsymbol{\tau}_x$ \cite{rangan2011generalized,Schniter_conv}:
\begin{flalign}
g_x(\boldsymbol{r},\boldsymbol{\tau}_r)=\frac{\boldsymbol{\gamma}}{\boldsymbol{\gamma + \tau_r}} \boldsymbol{r}  \label{eq:gx}\\
 g^{\prime}_x(\boldsymbol{r},\boldsymbol{\tau}_r)= \frac{\boldsymbol{\gamma}}{\boldsymbol{\gamma} + \boldsymbol{\tau}_r}. \label{eq:tau_x}
\end{flalign}

Similarly, in the case of the likelihood $p(\boldsymbol{y} \mathcal{\vert} \boldsymbol{x})$ given in \eqref{eq:py}, the max-sum and sum-product versions of $g_s(\boldsymbol{p},\boldsymbol{\tau}_p)$ yield the same updates for $\boldsymbol{s}$ and $\boldsymbol{\tau}_s$ \cite{rangan2011generalized,Schniter_conv}:

\begin{flalign}
g_s(\boldsymbol{p},\boldsymbol{\tau}_p) &= \frac{(\boldsymbol{p}/\boldsymbol{\tau}_p-\boldsymbol{y})}{(\sigma^2+1/\boldsymbol{\tau}_p)} \label{eq:gs} \\
g^{\prime}_s(\boldsymbol{p},\boldsymbol{\tau}_p)&=\frac{\sigma^{-2} }{\sigma^{-2} + \boldsymbol{\tau}_p}. \label{eq:tau_s}
\end{flalign}

We note that, in equations \eqref{eq:gx},\eqref{eq:tau_x},\eqref{eq:gs} and \eqref{eq:tau_s}, and for all equations in Table~\ref{tab:SMV_algorithm}, all vector squares, divisions and multiplications are taken element wise.

\begin{table}[h]
\centering
\begin{tabular}{|l r|}
\hline
Initialization & \\
$\boldsymbol{S} \leftarrow |\boldsymbol{A}|^2$ (component wise magnitude squared)  & (I1)\\
Initialize $\breve{\boldsymbol{\tau}}^0_x , \boldsymbol{\gamma}^0, (\sigma^2)^0 >0$ & (I2)\\
$\breve{\boldsymbol{s}}^{0},\breve{\boldsymbol{x}}^0 \leftarrow \boldsymbol{0}$  & (I3)\\
\hline
for $i=1,2,....,I_\text{max}$ & \\
\ \ \ Initialize $\boldsymbol{\tau}^{1}_x \leftarrow \breve{\boldsymbol{\tau}}^{i-1}_x , \hat{\boldsymbol{x}}^{1} \leftarrow \breve{\boldsymbol{x}}^{i-1} , \boldsymbol{s}^{1} \leftarrow \breve{\boldsymbol{s}}^{i-1}$ &\\
\ \ \ E-Step approximation & \\
\ \ \ for $k=1,2,....,K_\text{max}$ & \\
\ \ \ $ \ \ \ 1/\boldsymbol{\tau}_p^k \leftarrow \boldsymbol{S} \boldsymbol{\tau}_x^k $ &(A1)\\
\ \ \ $\ \ \ \boldsymbol{p}^k \leftarrow \boldsymbol{s}^{k-1} + \boldsymbol{\tau}_p^k \boldsymbol{A} \hat{\boldsymbol{x}}^k$ & (A2)\\
\ \ \ $\ \ \ \boldsymbol{\tau}_s^k \leftarrow \boldsymbol{\tau}_p^k g^{\prime}_s(\boldsymbol{p}^k,\boldsymbol{\tau}_p^k)$ & (A3)\\
\ \ \ $\ \ \ \boldsymbol{s}^k \leftarrow (1-\theta_s) \boldsymbol{s}^{k-1} + \theta_s g_s(\boldsymbol{p}^k,\boldsymbol{\tau}_p^k)$ & (A4)\\
\ \ \ $\ \ \ 1/\boldsymbol{\tau}_r^k \leftarrow \boldsymbol{S}^\top \boldsymbol{\tau}_s^k$ & (A5)\\
\ \ \ $\ \ \ \boldsymbol{r}^k \leftarrow \hat{\boldsymbol{x}}^k - \boldsymbol{\tau}_r^k \boldsymbol{A}^\top \boldsymbol{s}^k$ & (A6)\\
\ \ \ $\ \ \ \boldsymbol{\tau}_x^{k+1} \leftarrow \boldsymbol{\tau}_r^k g^{\prime}_x(\boldsymbol{r}^k,\boldsymbol{\tau}_r^k)$  & (A7)\\
\ \ \ $\ \ \ \hat{\boldsymbol{x}}^{k+1} \leftarrow (1-\theta_x)\hat{\boldsymbol{x}}^k + \theta_x g_x(\boldsymbol{r}^k,\boldsymbol{\tau}_r^k)$ & (A8)\\
\ \ \ \ \ \ if $\|\hat{\boldsymbol{x}}^{k+1}-\hat{\boldsymbol{x}}^k\|^2 / \|\hat{\boldsymbol{x}}^{k+1}\|^2 < \epsilon_\text{gamp}$ , break & (A9)\\
\ \ \ end for \%end of k loop & \\
\ \ \ $\breve{\boldsymbol{s}}^i \leftarrow {\boldsymbol{s}}^{k}$, $\breve{\boldsymbol{x}}^i \leftarrow \hat{\boldsymbol{x}}^{k+1}$ , $\breve{\boldsymbol{\tau}}^i_x \leftarrow \boldsymbol{\tau}^{k+1}_{x}$ &\\
\ \ \ M-Step & \\
\ \ \ $\boldsymbol{\gamma}^{i+1} \leftarrow |\breve{\boldsymbol{x}}^i|^2+\breve{\boldsymbol{\tau}}^i_x$ & (M1)\\
\ \ \ ${{(\sigma^{2})}^{i+1}} \leftarrow \frac{\| \boldsymbol{y} - \boldsymbol{A} \breve{\boldsymbol{x}}^i \|^2 + {(\sigma^2)}^i \sum_{n=1}^N \left( 1-  \frac{\breve{\tau}^i_{x_n}}{\gamma_n^i} \right)}{M}$  & (M2)\\
\ \ \ if $\|\breve{\boldsymbol{x}}^{i}-\breve{\boldsymbol{x}}^{i-1}\|^2 / \|\breve{\boldsymbol{x}}^{i}\|^2 < \epsilon_\text{em}$ , break & (M3)\\
end for \%end of i loop & \\
\hline
\end{tabular}
\caption{GGAMP-SBL algorithm}
\label{tab:SMV_algorithm}
\end{table}

In Table~\ref{tab:SMV_algorithm}, $K_\text{max}$ is the maximum allowed number of GAMP algorithm iterations, $\epsilon_\text{gamp}$ is the GAMP normalized tolerance parameter, $I_\text{max}$ is the maximum allowed number of EM iterations and $\epsilon_\text{em}$ is the EM normalized tolerance parameter.
Upon the convergence of GAMP algorithm based E-step, estimates for the mean $\hat{\boldsymbol{x}}$ and covariance diagonal $\boldsymbol{\tau}_x$ are obtained. These estimates can be used in the M-step of the algorithm, given by equation \eqref{eq:mstep2}. These estimates, along with the $\boldsymbol{s}$ vector estimate, are also used to initialize the E-step at the next EM iteration to accelerate the convergence of the overall algorithm.

Defining $\boldsymbol{S}$ as the component wise magnitude squared of $\boldsymbol{A}$, the complexity of the GGAMP-SBL algorithm is dominated by the E-step, which in turn (from Table~\ref{tab:SMV_algorithm}) is dominated by the matrix multiplications by $\boldsymbol{A},$ $\boldsymbol{A}^\top,$ $\boldsymbol{S}$ and $\boldsymbol{S}^\top$ at each iteration, implying that the computational cost of the algorithm is $\mathcal{O}(NM)$ operations per GAMP algorithm iteration  multiplied by the total number of GAMP algorithm iterations.  For large $M$, this is much smaller than $\mathcal{O} (N M^2)$, the complexity of standard SBL iteration.

In addition to the complexity of each iteration, for the proposed GGAMP-SBL algorithm to achieve faster runtimes it is important for GGAMP-SBL total number of iterations to not be too large, to the point where it over weighs the reduction in complexity per iteration, especially when heavier damping is used. We point out here that while SBL provides a one step exact solution for the E-step, GGAMP-SBL provides an approximate iterative solution. Based on that, the total number of SBL iterations is the number of EM iterations needed for convergence, while the total number of GGAMP-SBL iterations is based on the number of EM iterations it needs to converge and the number of E-step iterations for each EM iteration.
First we consider the number of EM iterations for both algorithms. As explained in Section \ref{convergence_total}, the E-step of GGAMP-SBL algorithm provides a good approximation of the true posterior \cite{ranganfixed}. In addition to that the number of EM iterations is not affected by damping, since damping only affects the number of iterations of GGAMP in the E-step, but it does not affect its outcome upon convergence. Based on these two points, we can expect the number of EM iterations for GGAMP-SBL to be generally in the same range as the original SBL algorithm. This is also shown in Section \ref{convergence_total} Figs.~\ref{cost1} and ~\ref{cost2}, where we can see the two cost functions being reduced to their minimum values using approximately the same number of EM iterations, even when heavier damping is used.
As for the GGAMP-SBL E-step iterations, because we are warm starting each E-step with $\boldsymbol{x}$ and $\boldsymbol{s}$ values from the previous EM iteration, it was found through numerical experiments that the number of required E-step iterations is reduced each time, to the point where the E-step converges to the required tolerance within 2-3 iterations towards the final EM iterations. When averaging the total number of E-step iterations over the number of EM iterations, it was found that for medium to large problem sizes the average number of E-step iterations was just a fraction of the measurements number $M$, even in the cases where heavier damping was used. Moreover, it was observed that the number of iterations required for the E-step to converge is independent of the problem size, which gives the algorithm a bigger advantage at larger problem sizes.
Finally, based on the complexity reduction per iteration and the total number of iterations required for GGAMP-SBL, we can expect it to have lower runtimes than SBL for medium to large problems, even when heavier damping is used. This runtime advantage is confirmed through numerical experiments in Section \ref{Numerical_Results}.


\subsection{GGAMP-SBL Convergence}
We now examine the convergence of the GGAMP-SBL algorithm. This involves two steps; the first step is to show that the approximate message passing algorithm employed in the E-step converges and the second step is to show that the overall EM algorithm, which consists of several E and M-steps, converges. For the second step, in addition to convergence of the E-step (first step), the accuracy of the resulting estimates is important. Therefore, in the second step of our convergence investigation, we use results from \cite{ranganfixed}, in addition to numerical results to show that the GGAMP-SBL's E and M steps are actually descending on the original SBL's cost function \eqref{eq:cost} at each EM iteration.

\subsubsection{Convergence of the E-step with Generic Transformations}
For the first step, we use the analysis from \cite{Schniter_conv} which shows that, in the case of generic $\boldsymbol{A}$,  the damped GGAMP algorithm is guaranteed to globally converge (to some values $\hat{\boldsymbol{x}}$ and $\boldsymbol{\tau}_x$) when sufficient damping is used. In particular, since $\boldsymbol{\gamma}^i$ is fixed in the E-step, the prior is Gaussian and so based on results in \cite{Schniter_conv}, starting with an initial estimate $\boldsymbol{\tau}_x \geq \boldsymbol{\gamma}^i$ the variance updates $\boldsymbol{\tau}_x$, $\boldsymbol{\tau}_s$, $\boldsymbol{\tau}_r$ and $\boldsymbol{\tau}_p$ will converge to a unique fixed point.
In addition,  any fixed point $(\boldsymbol{s}, \hat{\boldsymbol{x}})$ for GGAMP is globally stable if $\theta_s \theta_x ||\tilde{\boldsymbol{A}}||_2^2 < 1$, where the matrix $\tilde{\boldsymbol{A}}$ is defined as given below and is based on the fixed-point values of $\boldsymbol{\tau}_p$ and $\boldsymbol{\tau}_r$:
\begin{flalign*}
\tilde{\boldsymbol{A}} &:= \text{Diag}^{1/2}(\boldsymbol{\tau}_p \boldsymbol{q}_s) \boldsymbol{A} \ \text{Diag}^{1/2}(\boldsymbol{\tau}_r \boldsymbol{q}_x)\\
\boldsymbol{q}_s &=\frac{\sigma^{-2} }{\sigma^{-2} + \boldsymbol{\tau}_p} , \ \boldsymbol{q}_x= \frac{\boldsymbol{\gamma}}{\boldsymbol{\gamma} + \boldsymbol{\tau}_r}.
\end{flalign*}

While the result above establishes that the GGAMP algorithm is guaranteed to converge when sufficient amount of damping is used at each iteration, in practice we do not recommend building the matrix $\tilde{\boldsymbol{A}}$ at each EM iteration and calculating its spectral norm. Rather, we recommend choosing sufficiently small damping factors $\theta_x$ and $\theta_s$ and fixing them for all GGAMP-SBL iterations. For this purpose, the following result from \cite{Schniter_conv} for an i.i.d.-Gaussian prior $p(\boldsymbol{x})=\mathcal{N}(\boldsymbol{x};\boldsymbol{0},\gamma_x \boldsymbol{I})$ can provide some guidance on  choosing the damping factors. For the i.i.d.-Gaussian prior case, the damped GAMP algorithm is shown to converge if
\begin{flalign}
\label{eq:damp_thresh2} &\Omega(\theta_s,\theta_x) >  \|\boldsymbol{A}\|_2^2/\|\boldsymbol{A}\|_F^2,
\end{flalign}
where $\Omega(\theta_s,\theta_x)$ is defined as
\begin{flalign}
\label{eq:damp_thresh1} &\Omega(\theta_s,\theta_x) := \frac{2[(2-\theta_x)N+\theta_x M]}{\theta_x \theta_s M N}.
\end{flalign}
Experimentally, it was found that using a threshold $\Omega(\theta_s,\theta_x)$ that is 10\% larger than \eqref{eq:damp_thresh1} is sufficient for the GGAMP-SBL algorithm to converge in the scenarios we considered.

\subsubsection{GGAMP-SBL Convergence} \label{convergence_total}
The result above guarantees convergence of the E-step to some vectors $\hat{\boldsymbol{x}}$ and $\boldsymbol{\tau}_x$  but it does not provide information about the overall convergence of the EM algorithm to the desired SBL fixed points. This convergence depends on the quality of the mean $\hat{\boldsymbol{x}}$ and variance $\boldsymbol{\tau}_x$ computed by the GGAMP algorithm.
It has been shown that for an arbitrary $\boldsymbol{A}$ matrix, the fixed-point value of $\hat{\boldsymbol{x}}$ will equal the true mean given in \eqref{eq:mux} \cite{ranganfixed}. As for the variance updates, based on the state evolution in \cite{bayati2011dynamics}, the vector $\boldsymbol{\tau}_x$ will equal the true posterior variance vector, i.e., the diagonal of \eqref{eq:Sigmax}, in the case that $\boldsymbol{A}$ is large and i.i.d. Gaussian, but otherwise will only approximate the true quantity.

The approximation of $\boldsymbol{\tau}_x$ by the GGAMP algorithm in the E-step introduces an approximation in the GGAMP-SBL algorithm compared to the original EM-SBL algorithm. Fortunately, there is some flexibility in the EM algorithm in that the M-step need not be carried out to
minimize the objective stated in \eqref{eq:mstep} but it is sufficient to decrease the objective function as discussed in the generalized EM algorithm literature \cite{wu1983convergence,mclachlan2007algorithm}.
Given that the mean is estimated accurately, EM iterations may be tolerant to some error in the variance estimate.
Some flexibility in this regards can also be gleaned from the results in \cite{wipf2010iterative}, where it is shown how different iteratively reweighted algorithms correspond to a different choice in the variance. However, we have not been
able to prove rigorously that the GGAMP approximation will guarantee descent of the original cost function given in \eqref{eq:cost}.
%

\begin{figure*}[t!]
    \centering
    \begin{subfigure}[t]{0.5 \textwidth}
        \centering
        \includegraphics[height=1.2in]{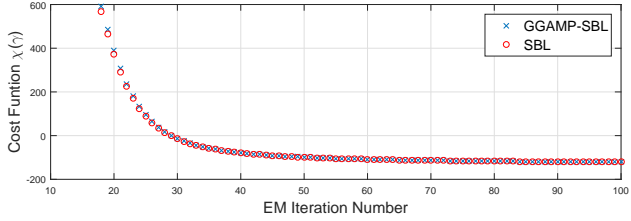}
        \caption{Cost functions for i.i.d.-Gaussian A} \label{cost1}
    \end{subfigure}%
    ~
    \begin{subfigure}[t]{0.5 \textwidth}
        \centering
        \includegraphics[height=1.2in]{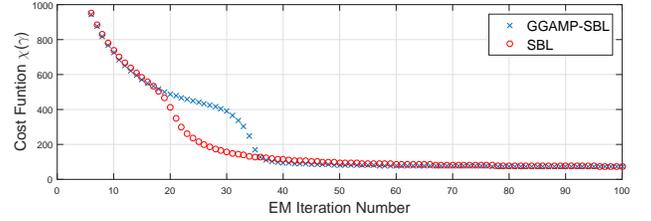}
        \caption{Cost functions for column correlated A} \label{cost2}
    \end{subfigure}
    \caption{Cost functions on SBL and GGAMP-SBL algorithms versus number of EM iterations}
    \label{cost}
\end{figure*}

Nevertheless, our numerical experiments  suggest that the GGAMP approximation has negligible effect on algorithm convergence and ability to recover sparse solutions.
We select two experiments  to illustrate the convergence behavior and demonstrate that the approximate variance estimates are sufficient to decrease SBL's cost function \eqref{eq:cost}. In both experiments $\boldsymbol{x}$ is drawn from a Bernoulli-Gaussian distribution with a non-zero probability $\lambda$ set to $0.2$, and we set $N=1000$ and $M=500$. Fig.~\ref{cost} shows a comparison between the original SBL and the GGAMP-SBL's cost functions at each EM iteration of the algorithms. $\boldsymbol{A}$ in Fig.~\ref{cost1} is i.i.d.-Gaussian, while in Fig.~\ref{cost2} it is a column correlated transformation matrix, which is constructed according to the description given in Section \ref{Numerical_Results}, with correlation coefficient $\rho=0.9$.

The cost functions in Fig.~\ref{cost1} and Fig.~\ref{cost2} show that, although we are using an approximate variance estimate to implement the M-step, the updates are decreasing the SBL's cost function at each iteration. As noted previously, it is not necessary for the M-step to provide the maximum cost function reduction, it is sufficient to provide some reduction in the cost function for the EM algorithm to be effective.
The cost function plots confirm this principle, since GGAMP-SBL eventually reaches the same minimal value as the original EM-SBL. While the two numerical experiments do not provide a guarantee that the overall GGAMP-SBL algorithm will converge, they suggest that the performance of the GGAMP-SBL algorithm often matches that of the original EM-SBL algorithm, which is supported by the more extensive numerical results in Section \ref{Numerical_Results}.

\section{GGAMP-TSBL for the MMV problem} \label{DGAMP_TSBL}
In this section, we apply the damped GAMP algorithm to the MMV empirical Bayesian approach to derive a low complexity algorithm for the MMV case as well.
Since the GAMP algorithm was originally derived for the SMV case using an SMV factor graph \cite{rangan2011generalized}, extending it to the MMV case requires some more effort and requires going back to the factor graphs that are the basis of the GAMP algorithm, making some adjustments, and then utilizing the GAMP algorithm.

Once again we use an empirical Bayesian approach with a GSM model, and we focus on the ML estimate of $\boldsymbol{\gamma}$. We assume a common sparsity profile between all measured vectors, and also account for the temporal correlation that might exist between the non-zero signal elements.
Previous Bayesian algorithms that have shown good recovery performance for the MMV problem include extensions of the SMV SBL algorithm, such as MSBL \cite{wipf2007empirical}, TSBL and TMSBL \cite{zhang2011sparse}. MSBL is a straightforward extension of SMV SBL, where no temporal correlation between non-zero elements is assumed, while TSBL and TMSBL account for temporal correlation. Even though the TMSBL algorithm has lower complexity compared to the TSBL algorithm, the algorithm still has complexity  of $\mathcal{O}(N M^2)$, which can limit its utility when the problem dimensions are large.
Other AMP based Bayesian algorithms have achieved linear complexity in the problem dimensions, like AMP-MMV \cite{AMP_MMV}. However AMP-MMV's robustness to generic $\boldsymbol{A}$ matrices is expected to be outperformed by an SBL based approach.
\subsection{MMV Model and Factor Graph}
The MMV model can be stated as:
\begin{flalign*}
&\boldsymbol{y}^{(t)}=\boldsymbol{Ax}^{(t)}+\boldsymbol{e}^{(t)} , \ \ t=1,2, ..., T,
\end{flalign*}
where we have $T$ measurement vectors $[\boldsymbol{y}^{(1)},\boldsymbol{y}^{(2)} ... , \boldsymbol{y}^{(T)}]$ with $\boldsymbol{y}^{(t)} \in \mathbb{R}^M.$ The objective is to recover $\boldsymbol{X} = [\boldsymbol{x}^{(t)},\boldsymbol{x}^{(2)} ... , \boldsymbol{x}^{(T)}]$ with  $\boldsymbol{x}^{(t)} \in \mathbb{R}^N$, where in addition to the vectors $\boldsymbol{x}^{(t)}$ being sparse, they share the same sparsity profile. Similar to the SMV case, $\boldsymbol{A} \in \mathbb{R}^{M \times N}$ is known, and $[\boldsymbol{e}^{(1)},\boldsymbol{e}^{(2)} ... , \boldsymbol{e}^{(T)}]$ is a sequence  of i.i.d. noise vectors modeled as $ \boldsymbol{e}^{(t)} \sim \mathcal{N} (0,\sigma^2 \boldsymbol{I})$.
This model can be restated as:
\begin{flalign*}
&\boldsymbol{\bar{y}=D(A)\bar{x}+\bar{e}},
\end{flalign*}
where $\boldsymbol{\bar{y}} \triangleq [\boldsymbol{y}^{(1)^\top},\boldsymbol{y}^{(2)^\top} ... , \boldsymbol{y}^{(T)^\top}]^\top,$ $\boldsymbol{\bar{x}} \triangleq [\boldsymbol{x}^{(1)^\top},\boldsymbol{x}^{(2)^\top} ... , \boldsymbol{x}^{(T)^\top}]^\top,$ $\boldsymbol{\bar{e}} \triangleq [\boldsymbol{e}^{(1)^\top},\boldsymbol{e}^{(2)^\top} ... , \boldsymbol{e}^{(T)^\top}]^\top$ and $\boldsymbol{D(A)}$ is a block-diagonal matrix constructed from $T$ replicas of $\boldsymbol{A}$.

\begin{figure}[h]
\centering
    \includegraphics[scale=.6]{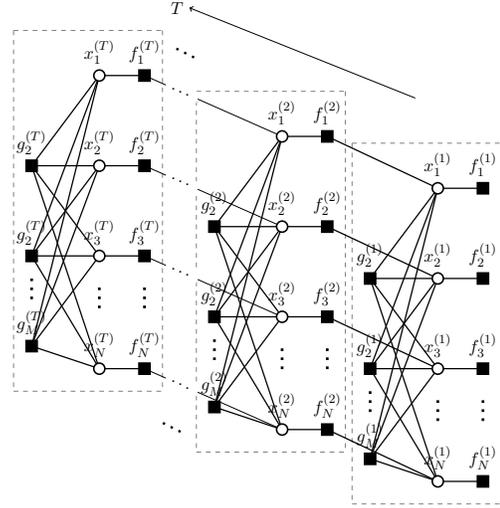}
\caption{GGAMP-TSBL factor graph}
\label{MMV_Factor_graph}
\end{figure}

The posterior distribution of $\boldsymbol{\bar{x}}$ is given by:
\begin{flalign*}
p(\boldsymbol{\bar{x} \mathbin{\vert} \bar{y}}) \  &\propto \
\prod_{t=1}^T \left[
\prod_{m=1}^{M}{p(y_m^{(t)} \mathbin{\vert} \boldsymbol{x}^{(t)} )}
\prod_{n=1}^{N}{p(x_n^{(t)} \mathbin{\vert} x_n^{(t-1)} )}\right],
\end{flalign*}
where
\begin{flalign*}
p(y_m^{(t)} \mathbin{\vert} \boldsymbol{x}^{(t)}) &= \mathcal{N}(y_m^{(t)};\boldsymbol{a}_{m,.}^\top \boldsymbol{x}^{(t)},\sigma^2),
\end{flalign*}
where $\boldsymbol{a}_{m,.}^\top$ is the $m^{\text{th}}$ row of the matrix $\boldsymbol{A}$. Similar to the previous work in \cite{zhang2010sparse,prasad2014joint,AMP_MMV}, we use an AR(1) process to model the correlation between $x_n^{(t)}$ and $x_n^{(t-1)},$ i.e.,
\begin{flalign*}
x_n^{(t)}&=\beta x_n^{(t-1)} + \sqrt{1-\beta^2} v_n^{(t)}\\
p(x_n^{(t)} \mathbin{\vert} x_n^{(t-1)}) &= \mathcal{N}(x_n^{(t)};\beta x_n^{(t-1)},(1-\beta^2) \gamma_n), \ t>1\\
p(x_n^{(1)}) &= \mathcal{N}(x_n^{(1)};0,\gamma_n),
\end{flalign*}
where $\beta \in (-1,1)$ is the temporal correlation coefficient and $v_n^{(t)} \sim \mathcal{N} (0,\gamma_n)$. Following an empirical Bayesian approach similar to the one proposed for the SMV case, the hyperparameter vector $\boldsymbol{\gamma}$ is then learned from the measurements using the EM algorithm. The EM algorithm can also be used to learn the correlation coefficient $\beta$ and the noise variance $\sigma^2$.
Based on these assumptions we use the sum-product algorithm \cite{kschischang2001factor} to construct the factor graph in Fig.~\ref{MMV_Factor_graph}, and derive the MMV algorithm GGAMP-TSBL. In the MMV factor graph, the factors are $g_m^{(t)}(\boldsymbol{x})=  p(y_m^{(t)} \mathbin{\vert} \boldsymbol{x}^{(t)})$, $f_n^{(t)}(x_n^{(t)})= p(x_n^{(t)} \mathbin{\vert} x_n^{(t-1)})$ for $t>1$ and $f_n^{(1)}(x_n^{(1)})=p(x_n^{(1)})$.

\begin{table}[t!]
\centering
\resizebox{\columnwidth}{!}{%
\begin{tabular}{|l r|}
\hline
Definitions & \\
$F(\boldsymbol{r}^{k(t)},\boldsymbol{\tau}_r^{k(t)})=\frac{\frac{\boldsymbol{r}^{k(t)}}{\boldsymbol{\tau}_r^{k(t)}}+\frac{\boldsymbol{\eta}^{k(t)}}{\boldsymbol{\psi}^{k(t)}}+\frac{\boldsymbol{\theta}^{k(t)}}{\boldsymbol{\phi}^{k(t)}}}{\frac{1}{\boldsymbol{\tau}_r^{k(t)}}+\frac{1}{\boldsymbol{\psi}^{k(t)}}+\frac{1}{\boldsymbol{\phi}^{k(t)}}}$ & (D1)\\
$G(\boldsymbol{r}^{k(t)},\boldsymbol{\tau}_r^{k(t)})=\frac{1}{\frac{1}{\boldsymbol{\tau}_r^{k(t)}}+\frac{1}{\boldsymbol{\psi}^{k(t)}}+\frac{1}{\boldsymbol{\phi}^{k(t)}}}$ & (D2)\\
\hline
Initialization & \\
$\boldsymbol{S} \leftarrow |\boldsymbol{A}|^2$ (component wise magnitude squared)  & (N1)\\
Initialize $ \forall t : \breve{\boldsymbol{\tau}}^{0(t)}_x, \boldsymbol{\gamma}^0 >0$,
$\breve{\boldsymbol{s}}^{0(t)} \leftarrow \boldsymbol{0} \ \text{and} \ \breve{\boldsymbol{x}}^{0(t)} \leftarrow \boldsymbol{0}$  & (N3)\\
\hline
for $i=1,2,....,I_\text{max}$ & \\
\ \ \ Initialize $\forall t :\boldsymbol{\tau}^{1(t)}_x \leftarrow \breve{\boldsymbol{\tau}}^{i-1(t)}_x , \hat{\boldsymbol{x}}^{1(t)} \leftarrow \breve{\boldsymbol{x}}^{i-1(t)},$ &\\
$ {\boldsymbol{s}}^{1(t)} \leftarrow \breve{\boldsymbol{s}}^{i-1(t)}$ &\\
\ \ \ E-Step approximation & \\
\ \ \ for $k=1,2,....,K_\text{max}$ & \\
\ \ \ $\ \ \boldsymbol{\eta}^{k(1)} \leftarrow 0$ & (E1)\\
\ \ \ $\ \ \boldsymbol{\psi}^{k(1)} \leftarrow \boldsymbol{\gamma}^i$ & (E2)\\
\ \ \ $\ \ \text{for } t=2:T$ & \\
\ \ \ $\ \ \ \ \boldsymbol{\eta}^{k(t)} \leftarrow \beta \left( \frac{\boldsymbol{r}^{k(t-1)}}{\boldsymbol{\tau}_r^{k(t-1)}}+\frac{\boldsymbol{\eta}^{k(t-1)}}{\boldsymbol{\psi}^{k(t-1)}}\right) \left(\frac{\boldsymbol{\psi}^{k(t-1)} \boldsymbol{\tau}_r^{k(t-1)}}{\boldsymbol{\psi}^{k(t-1)}+\boldsymbol{\tau}_r^{k(t-1)}} \right)$ & (E3)\\
\ \ \ $\ \ \ \ \boldsymbol{\psi}^{k(t)} \leftarrow \beta^2 \left(\frac{\boldsymbol{\psi}^{k(t-1)} \boldsymbol{\tau}_r^{k(t-1)}}{\boldsymbol{\psi}^{k(t-1)}+\boldsymbol{\tau}_r^{k(t-1)}} \right)+(1-\beta^2) \boldsymbol{\gamma}^i$ & (E4)\\
\ \ \ $\ \ \text{end for}$ \%end of t loop &\\
\ \ \ $\ \ \text{for } t=1:T$ & \\
\ \ \ $\ \ \ \ 1/\boldsymbol{\tau}_p^{k(t)} \leftarrow \boldsymbol{S} \boldsymbol{\tau}_x^{k(t)}$ & (E5)\\
\ \ \ $\ \ \ \ \boldsymbol{p}^{k(t)} \leftarrow \boldsymbol{s}^{k-1(t)} + \boldsymbol{\tau}_p^{k(t)} \boldsymbol{A} \hat{\boldsymbol{x}}^{k(t)}$  & (E6)\\
\ \ \ $\ \ \ \ \boldsymbol{\tau}_s^{k(t)} \leftarrow \frac{\sigma^{-2} \boldsymbol{\tau}_p^{k(t)}}{\sigma^{-2} + \boldsymbol{\tau}_p^{k(t)}}$& (E7)\\
\ \ \ $\ \ \ \ \boldsymbol{s}^{k(t)} \leftarrow (1-\theta_s)\boldsymbol{s}^{k-1(t)} + \theta_s \frac{\left( \frac{\boldsymbol{p}^{k(t)}}{\boldsymbol{\tau}_p^{k(t)}}-\boldsymbol{y}^{(t)}\right) }{(\sigma^2+1/\boldsymbol{\tau}_p^{k(t)})}$ & (E8)\\
\ \ \ $\ \ \ \ 1/\boldsymbol{\tau}_r^{k(t)} \leftarrow \boldsymbol{S}^\top \boldsymbol{\tau}_s^{k(t)}$  & (E9)\\
\ \ \ $\ \ \ \ \boldsymbol{r}^{k(t)} \leftarrow \hat{\boldsymbol{x}}^{(t)} - \boldsymbol{\tau}_r^{k(t)} \boldsymbol{A}^\top \boldsymbol{s}^{k(t)}$ & (E10)\\
\ \ \ $\ \ \ \ \boldsymbol{\tau}_x^{k+1(t)} \leftarrow G(\boldsymbol{r}^{k(t)},\boldsymbol{\tau}_r^{k(t)})$ & (E11)\\
\ \ \ $\ \ \ \ \hat{\boldsymbol{x}}^{k+1(t)} \leftarrow (1-\theta_x)\hat{\boldsymbol{x}}^{k(t)} + \theta_x F_n(\boldsymbol{r}^{k(t)},\boldsymbol{\tau}_r^{k(t)})$ & (E12)\\
\ \ \ $\ \ \text{end for}$ \%end of t loop &\\
\ \ \ $\ \ \text{for } t=T-1:1$ & \\
\ \ \ $\ \ \ \ \boldsymbol{\theta}^{k(t)} \leftarrow \frac{1}{\beta} \left( \frac{\boldsymbol{r}^{k(t+1)}}{\boldsymbol{\tau}_r^{k(t+1)}}+\frac{\boldsymbol{\theta}^{k(t+1)}}{\boldsymbol{\phi}^{k(t+1)}}\right)\left(\frac{\boldsymbol{\phi}^{k(t+1)} \boldsymbol{\tau}_r^{k(t+1)}}{\boldsymbol{\theta}^{k(t+1)}+\boldsymbol{\tau}_r^{k(t+1)}} \right)$ & (E13)\\
\ \ \ $\ \ \ \ \boldsymbol{\phi}^{k(t)} \leftarrow \frac{1}{\beta^2}\left( \frac{\boldsymbol{\phi}^{k(t+1)} \boldsymbol{\tau}_r^{k(t+1)}}{\boldsymbol{\phi}^{k(t+1)}+\boldsymbol{\tau}_r^{k(t+1)}}+(1-\beta^2)\boldsymbol{\gamma}^i \right)$ & (E14)\\
\ \ \ $\ \ \text{end for}$ \%end of t loop & \\
\ \ \ \ \ \ if $\frac{1}{T}\sum_{t=1}^T \left( \frac{\|\hat{\boldsymbol{x}}^{k+1(t)}- \hat{\boldsymbol{x}}^{k(t)}\|^2} {\|\hat{\boldsymbol{x}}^{k+1(t)}\|^2} \right) < \epsilon_\text{gamp}$ , break & (E15)\\
\ \ \ end for \%end of k loop &\\
\ \ \ $\forall t, \breve{\boldsymbol{s}}^{i(t)} \leftarrow {\boldsymbol{s}}^{k+1(t)}, \breve{\boldsymbol{x}}^{i(t)} \leftarrow \hat{\boldsymbol{x}}^{k+1(t)}$ , $\breve{\boldsymbol{\tau}}^{i(t)}_x \leftarrow \boldsymbol{\tau}^{k+1(t)}_{x}$ &\\
\ \ \ M-step & \\
\ \ \ $\gamma_n^{i+1}=\frac{1}{T} \biggl[ |\breve{x}_n^{i(1)}|^2 + \breve{\tau}_{x_n}^{i(1)} +\sum_{t=2}^T \frac{|\breve{x}_n^{{i(t)}}|^2 + \breve{\tau}_{x_n}^{i(t)}}{1-\beta^2}$ &\\
\ \ \ \ \ \ \ \ \ $+ \frac{\beta}{1-\beta^2} \sum_{t=2}^T \left(|\breve{x}_n^{i(t-1)}|^2 + \breve{\tau}_{x_n}^{i(t-1)}\right)$ &\\
\ \ \ \ \ \ \ \ \ $- \frac{2 \beta}{1-\beta^2} \sum_{t=2}^T \left(\breve{x}_n^{i(t)} \breve{x}_n^{i(t-1)} + \beta \breve{x}_{n}^{i(t-1)}\right) \biggr ] $ & (U1)\\
\ \ \ if $\frac{1}{T}\sum_{t=1}^T \left( \frac{\|\breve{\boldsymbol{x}}^{i(t)}-\breve{\boldsymbol{x}}^{i-1(t)}\|^2}  {\|\breve{\boldsymbol{x}}^{i(t)}\|^2} \right) < \epsilon_\text{em} $ , break & (U2)\\
end for \%end of i loop & \\
\hline
\end{tabular}
}
\caption{GGAMP-TSBL algorithm}
\label{tab:MMV_algorithm}
\end{table}

\subsection{GGAMP-TSBL Message Phases and Scheduling (E-Step)}
\begin{figure}[h]
\centering
    \includegraphics[scale=.5]{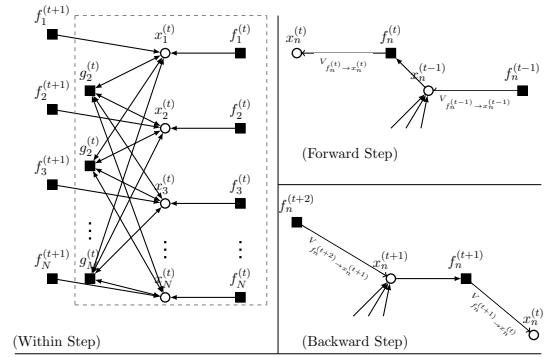}
\caption{Message passing phases for GGAMP-TSBL} \label{Message_Steps}
\end{figure}

Due to the similarities between the factor graph for each time frame of the MMV model and the factor graph of the SMV model, we will use the algorithm in Table~\ref{tab:SMV_algorithm} as a building block and extend it to the MMV case. We divide the message updates into three steps as shown in Fig.~\ref{Message_Steps}.

\small
\begin{table*}[b]
\HRule[1pt]
\begin{flalign}
\gamma_n^{i+1} &=\frac{1}{T} \left[ |\hat{x}_n^{(1)}|^2 + \tau_{x_n}^{(1)}+ \frac{1}{1-\beta^2}\sum_{t=2}^T |\hat{x}_n^{(1)}|^2 + \tau_{x_n}^{(t)} + \beta (|\hat{x}_n^{(t-1)}|^2 + \tau_{x_n}^{(t-1)}) - 2 \beta (\hat{x}_n^{(t)} \hat{x}_n^{(t-1)} + \beta \tau_{x_n}^{(t-1)}) \right]. \label{eq:MMV_M-step}
\end{flalign}
\end{table*}
\normalsize

For each time frame the ``within`` step in Fig.~\ref{Message_Steps} is very similar to the SMV GAMP iteration, with the only difference being that each $x_n^{(t)}$ is connected to the factor nodes $f_n^{(t)}$ and $f_n^{(t+1)}$, while it is connected to one factor node in the SMV case. This difference is reflected in the calculation of the output function $g_x$ and therefore in finding the mean and variance estimates for $\bar{\boldsymbol{x}}$. The details of finding $g_x$ and therefore the update equations for $\boldsymbol{\tau}_x^{(t)}$ and $\hat{\boldsymbol{x}}^{(t)}$ are shown in Appendix A. The input function $g_s$ is the same as \eqref{eq:gs}, and the update equations for $\boldsymbol{\tau}_s^{(t)}$ and $\boldsymbol{s}^{(t)}$ are the same as (A3) and (A4) from Table~\ref{tab:SMV_algorithm}, because an AWGN model is assumed for the noise.
The second type of updates are passing messages forward in time from $x_n^{(t-1)}$ to $x_n^{(t)}$ through $f_n^{(t)}$.
And the final type of updates is passing messages backward in time from $x_n^{(t+1)}$ to $x_n^{(t)}$ through $f_n^{(t)}$. The details for finding the ``forward`` and ``backward`` message passing steps are also shown in Appendix A.\\
We schedule the messages by moving forward in time first, where we run the ``forward`` step starting at $t=1$ all the way to $t=T$. We then perform the ``within`` step for all time frames, this step updates $\boldsymbol{r}^{(t)}$,$\boldsymbol{\tau}_r^{(t)}$, $\boldsymbol{x}^{(t)}$ and $\boldsymbol{\tau}_x^{(t)}$ that are needed for the ``forward`` and ``backward`` message passing steps. Finally we pass the messages backward in time using the ``backward`` step, starting at $t=T$ and ending at $t=1$.
Based on this message schedule, the GAMP algorithm E-step computation is summarized in Table~\ref{tab:MMV_algorithm}.
In Table~\ref{tab:MMV_algorithm} we use the unparenthesized superscript to indicate the iteration index, while the parenthesized superscript indicates the time frame index. Similar to Table~\ref{tab:SMV_algorithm}, $K_\text{max}$ is the maximum allowed number of GAMP iterations, $\epsilon_\text{gamp}$ is the GAMP normalized tolerance parameter, $I_\text{max}$ is the maximum allowed number of EM iterations and $\epsilon_\text{em}$ is the EM normalized tolerance parameter. In Table~\ref{tab:MMV_algorithm} all vector squares, divisions and multiplications are taken element wise.

The algorithm proposed can be considered an extension of the previously proposed AMP TSBL algorithm in \cite{AMPSBL}. The extension to GGAMP-TSBL includes removing the averaging of the matrix $\boldsymbol{A}$ in the derivation of the algorithm, and it includes introducing the same damping strategy used in the SMV case to improve convergence.
The complexity of the GGAMP-TSBL algorithm is also dominated by the E-step which in turn is dominated by matrix multiplications by $\boldsymbol{A},$ $\boldsymbol{A}^\top,$ $\boldsymbol{S}$ and $\boldsymbol{S}^\top,$ implying that the computational cost is $\mathcal{O}(MN)$ flops per iteration per frame. Therefore the complexity of the proposed algorithm is $\mathcal{O}(TMN)$ multiplied by the total number of GAMP algorithm iterations.

\subsection{GGAMP-TSBL M-Step}
Upon the convergence of the E-step, the M-step learns $\boldsymbol{\gamma}$ from the data by treating $\boldsymbol{x}$ as a hidden variable and then maximizing $\mathbb{E}_{\boldsymbol{\bar{x}} \mathcal{\vert} \boldsymbol{\bar{y}} ; \boldsymbol{\gamma}^{i}, \sigma^2, \beta} [\log p(\boldsymbol{\bar{y}}, \boldsymbol{\bar{x}} , \boldsymbol{\gamma}; \sigma^2 , \beta)]$.

\begin{flalign}
&\boldsymbol{\gamma}^{i+1}= \argmin_{\boldsymbol\gamma} \mathbb{E}_{\boldsymbol{\bar{x}} \mathcal{\vert} \boldsymbol{\bar{y}} ; \boldsymbol{\gamma}^{i}, \sigma^2, \beta} [-\log p(\boldsymbol{\bar{y}}, \boldsymbol{\bar{x}} , \boldsymbol{\gamma}; \sigma^2, \beta)]. \nonumber
\end{flalign}
\normalsize

The derivation of $\gamma_n^{i+1}$ M-step update follows the same steps as the SMV case. The derivation is omitted here due to space limitation, and $\gamma_n^{i+1}$ update is given in \eqref{eq:MMV_M-step} at the bottom of this page. We note here that the M-step $\boldsymbol{\gamma}$ learning rule in \eqref{eq:MMV_M-step} is the same as the one derived in \cite{prasad2014joint}. Both algorithms use the same AR(1) model for $\boldsymbol{x}^{(t)}$, but they differ in the implementation of the E-step.
In the case that the correlation coefficient $\beta$ or the noise variance $\sigma^2$ are unknown, the EM algorithm can be used to estimate their values as well.

\section{Numerical Results} \label{Numerical_Results}
In this section we present  a numerical study  to illustrate the performance and complexity of the proposed GGAMP-SBL and GGAMP-TSBL algorithms.
The performance and complexity were studied through two metrics. The first metric studies the ability of the algorithm to recover $\boldsymbol{x}$, for which we use the normalized mean squared error NMSE in the SMV case:
\begin{flalign*}
&\text{NMSE} \triangleq  \| \boldsymbol{\hat{x}} - \boldsymbol{x} \|^2 /  \| \boldsymbol{x} \|^2 ,
\end{flalign*}
and the time-averaged normalized mean squared error TNMSE in the MMV case:
\begin{flalign*}
&\text{TNMSE} \triangleq  \frac{1}{T} \sum_{t=1}^T \| \hat{\boldsymbol{x}}^{(t)} - \boldsymbol{x}^{(t)} \|^2 / \| \boldsymbol{x}^{(t)} \|^2 .
\end{flalign*}
The second metric studies the complexity of the algorithm by tracking the time the algorithm requires to compute the final estimate $\hat{\boldsymbol{x}}$. We measure the time in seconds. While the absolute runtime could vary if the same experiments were to be run on a different machine,  the runtimes of the algorithms of interest in relationship to each other is a good estimate of the relative computational complexity.

Several types of non-i.i.d.-Gaussian matrix were used to explore the robustness of the proposed algorithms relative to the standard SBL and TMSBL. The four different types of matrices are similar to the ones previously used in \cite{MADGAMP} and are described as follows:

-Column correlated matrices: The rows of $\boldsymbol{A}$ are independent zero-mean Gaussian Markov processes with the following correlation coefficient $\rho = \mathbb{E} \{\boldsymbol{a}^\top_{.,n} \boldsymbol{a}_{.,n+1}\}/\mathbb{E} \{|\boldsymbol{a}_{.,n}|^2\}$, where $\boldsymbol{a}_{.,n}$ is the $n^{\text{th}}$ column of $\boldsymbol{A}$. In the experiments the correlation coefficient $\rho$ is used as the measure of deviation from the i.i.d.-Gaussian matrix.

-Low rank product matrices: We construct a rank deficient $\boldsymbol{A}$ by $\boldsymbol{A}=\frac{1}{N} \boldsymbol{H G}$ with $\boldsymbol{H} \in \mathbb{R}^{M \times R}$, $\boldsymbol{G} \in \mathbb{R}^{R \times N}$ and $R < M$. The entries of $\boldsymbol{H}$ and $\boldsymbol{G}$ are i.i.d.-Gaussian with zero mean and unit variance. The rank ratio $R/N$ is used as the measure of deviation from the i.i.d.-Gaussian matrix.

-Ill conditioned matrices: we construct $\boldsymbol{A}$ with a condition number $\kappa > 1$ as follows. $\boldsymbol{A}=\boldsymbol{U \Sigma V}^\top$, where $\boldsymbol{U}$ and $\boldsymbol{V}^\top$ are the left and right singular vector matrices of an i.i.d.-Gaussian matrix, and $\boldsymbol{\Sigma}$ is a singular value matrix with $\Sigma_{i,i}/\Sigma_{i+1,i+1}=\kappa^{1/{(M-1)}}$ for $i=1,2,....,M-1$. The condition number $\kappa$ is used as the measure of deviation from the i.i.d.-Gaussian matrix.

-Non-zero mean matrices: The elements of $\boldsymbol{A}$ are $a_{m,n} \sim \mathcal{N} (\mu,\frac{1}{N})$. The mean $\mu$ is used as a measure of deviation from the zero-mean i.i.d.-Gaussian matrix. It is worth noting that in the case of non-zero mean $\boldsymbol{A},$ convergence of the GGAMP-SBL is not enhanced by damping but more by the mean removal procedure explained in \cite{MADGAMP}. We include it in the implementation of our algorithm, and we include it in the numerical results to make the study more inclusive of different types of generic $\boldsymbol{A}$ matrices.\\
Although we have provided an estimation procedure, based on the EM algorithm, for the noise variance $\sigma^2$ in \eqref{eq:sigma2}, in all experiments we assume that the noise variance $\sigma^2$ is known. We also found  that the SBL algorithm does not necessarily have the best performance when the exact $\sigma^2$ is used, and in our case, it was empirically found that using an estimate $\hat{\sigma}^2=3 \sigma^2$ yields better results. Therefore $\hat{\sigma}^2$ is used for SBL, TMSBL, GGAMP-SBL and GGAMP-TSBL throughout our experiments.
\subsection{SMV GGAMP-SBL Numerical Results}
In this section we compare the proposed SMV algorithm (GGAMP-SBL) against the original SBL and against two AMP algorithms that have shown improvement in robustness over the original AMP/GAMP, namely the SwAMP algorithm \cite{swept} and the MADGAMP algorithm \cite{MADGAMP}. As a performance benchmark, we use a lower bound on the achievable NMSE which is similar to the one in \cite{MADGAMP}. The bound is found using a ``genie`` that knows the support of the sparse vector $\boldsymbol{x}$. Based on the known support, $\boldsymbol{\bar{A}}$ is constructed from the columns of $\boldsymbol{A}$ corresponding to non-zero elements of $\boldsymbol{x}$, and an MMSE solution using $\boldsymbol{\bar{A}}$ is computed.
\begin{flalign*}
&\hat{\boldsymbol{x}}=\boldsymbol{\bar{A}}^\top(\boldsymbol{\bar{A}} \boldsymbol{\bar{A}}^\top + \sigma^2 \boldsymbol{I})^{-1} \boldsymbol{y}.
\end{flalign*}
In all SMV experiments, $\boldsymbol{x}$ had exactly $K$ non-zero elements in random locations, and the nonzero entires were drawn independently from a zero-mean unit-variance Gaussian distribution. In accordance with the model \eqref{eq:SMVmodel}, an AWGN channel was used with the SNR defined by:
\begin{flalign*}
&\text{SNR} \triangleq \mathbb{E} \{ \| \boldsymbol{Ax}\|^2 \} / \mathbb{E} \{ \| \boldsymbol{y - Ax} \|^2 \}.
\end{flalign*}

\begin{figure}[tb!]
\centering
\includegraphics[width= \columnwidth]{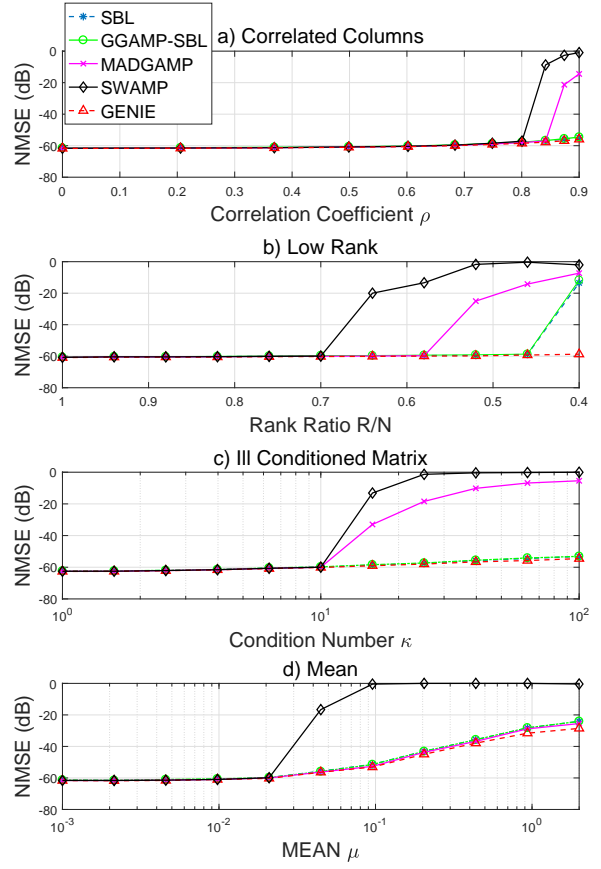}
\caption{NMSE comparison of SMV algorithms under non-i.i.d.-Gaussian $\boldsymbol{A}$ matrices with SNR=60dB} \label{SMV_60dB_MSE}
\centering
\end{figure}

\begin{figure}[tb!]
\centering
\includegraphics[width= \columnwidth]{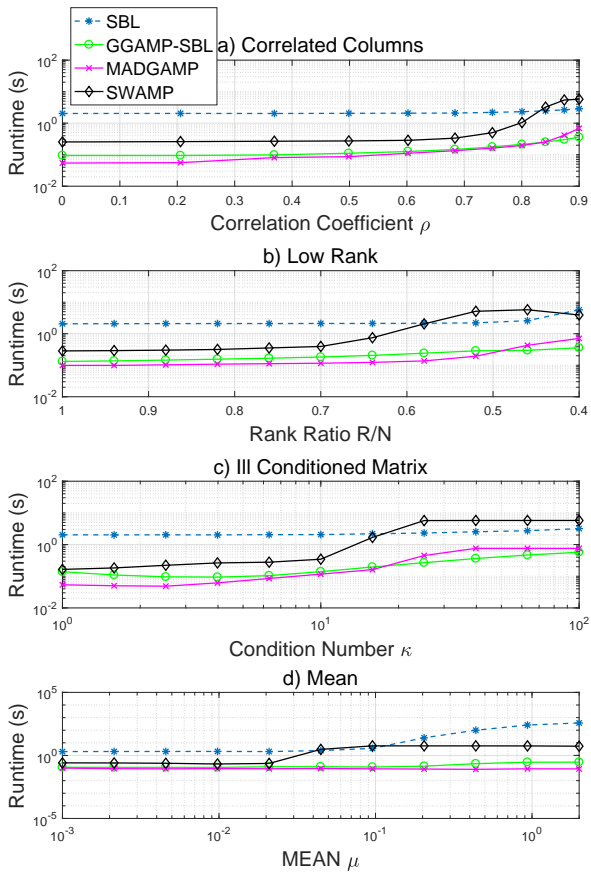}
\caption{Runtime comparison of SMV algorithms under non-i.i.d.-Gaussian $\boldsymbol{A}$ matrices with SNR=60dB} \label{SMV_60dB_TIME}
\end{figure}

\subsubsection{Robustness to generic matrices at high SNR}
The first experiment investigates the robustness of the proposed algorithm to generic $\boldsymbol{A}$ matrices. It compares the algorithms of interest using the four types of matrices mentioned above, over a range of deviation from the i.i.d.-Gaussian case. For each matrix type, we start with an i.i.d.-Gaussian $\boldsymbol{A}$ and increase the deviation over 11 steps. We monitor how much deviation the different algorithms can tolerate, before we start seeing significant performance degradation compared to the ``genie`` bound. The vector $\boldsymbol{x}$ was drawn from a Bernoulli-Gaussian distribution with non-zero probability $\lambda = 0.2$, with $N=1000$, $M=500$ and $\text{SNR}=60$dB.

The NMSE results in Fig.~\ref{SMV_60dB_MSE} show that the performance of GGAMP-SBL was able to match that of the original SBL even for $\boldsymbol{A}$ matrices with the most deviation from the i.i.d.-Gaussian case. Both algorithms nearly achieved the bound in most cases, with the exception when the matrix is low rank with a rank ratio less than $0.45$ where both algorithms fail to achieve the bound. This supports the evidence we provided before for the convergence of the GGAMP-SBL algorithm, which predicted its ability to match the performance of the original SBL.
As for other AMP implementations, despite the improvement in robustness they provide over traditional AMP/GAMP, they cannot guarantee convergence beyond a certain point, and their robustness is surpassed by GGAMP-SBL in most cases. The only exception is when $\boldsymbol{A}$ is non-zero mean, where the GGAMP-SBL and the MADGAMP algorithms share similar performance. This is due to the fact that both algorithms use mean removal to transform the problem into a zero-mean equivalent problem, which both algorithms can handle well.\\
The complexity of the GGAMP-SBL algorithm is studied in Fig.~\ref{SMV_60dB_TIME}. The figure shows how the GGAMP-SBL was able to reduce the complexity compared to the original SBL implementation. It also shows that even when the algorithm is slowed down by heavier damping, the algorithm still has faster runtimes than the original SBL.
\begin{figure}[tbh!]
\centering
\includegraphics[width= \columnwidth]{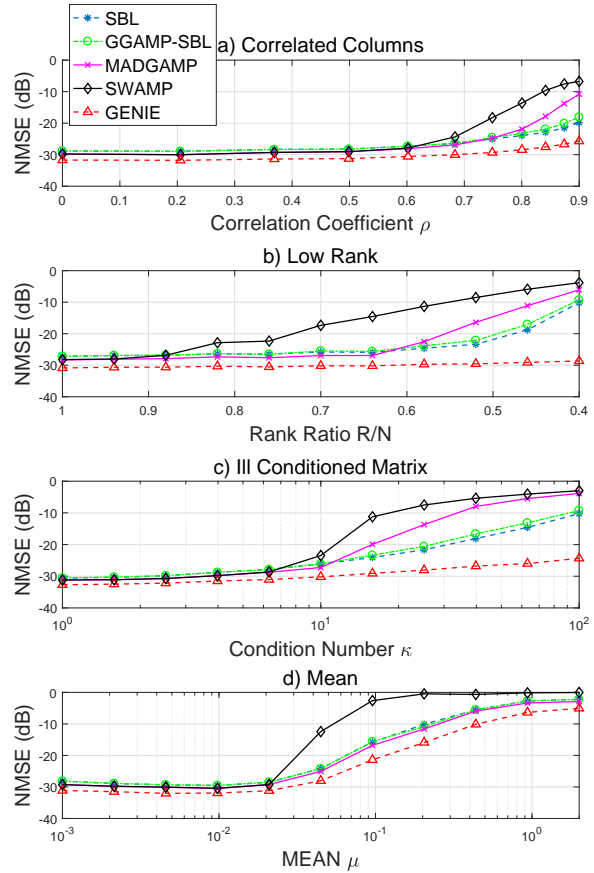}
\caption{NMSE comparison of SMV algorithms under non-i.i.d.-Gaussian $\boldsymbol{A}$ matrices with SNR=30dB} \label{SMV_30dB_MSE}
\end{figure}

\begin{figure}[tbh!]
\centering
\includegraphics[width= \columnwidth]{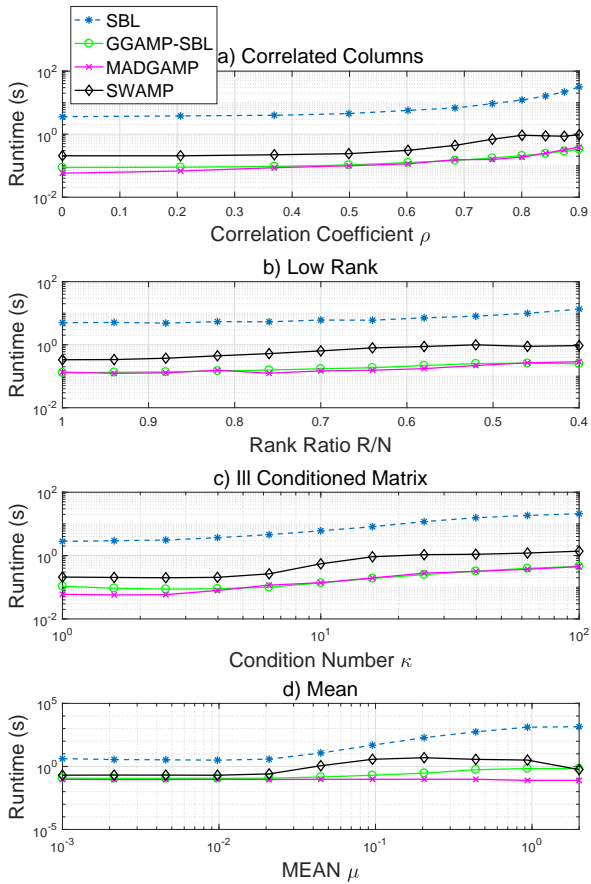}
\caption{Runtime comparison of SMV algorithms under non-i.i.d.-Gaussian $\boldsymbol{A}$ matrices with SNR=30dB} \label{SMV_30dB_TIME}
\end{figure}
\subsubsection{Robustness to generic matrices at lower SNR}
In this experiment we  examine the performance and complexity of the proposed algorithm at a lower SNR setting than the previous experiment. We lower the SNR to 30dB and collect the same data points as in the previous experiment.
The results in Fig.~\ref{SMV_30dB_MSE} show that the performance of the GGAMP-SBL algorithm is still generally matching that of the original SBL algorithm with slight degradation. The MADGAMP algorithm provides slightly better performance than both SBL algorithms when the deviation from the i.i.d.-sub-Gaussian case is not too large. This can be due to the fact that we choose to run the MADGAMP algorithm with exact knowledge of the data model rather than learn the model parameters, while both SBL algorithms have information about the noise variance only. As the deviation in $\boldsymbol{A}$ increases, GGAMP-SBL's performance surpasses MADGAMP and SWAMP algorithms, providing better robustness at lower SNR.\\
On the complexity side, we see from Fig.~\ref{SMV_30dB_TIME} that the GGAMP-SBL continues to have reduced complexity compared to the original SBL.


\begin{figure}
\begin{minipage}{\columnwidth}
\centering
\includegraphics[width=2.8in]{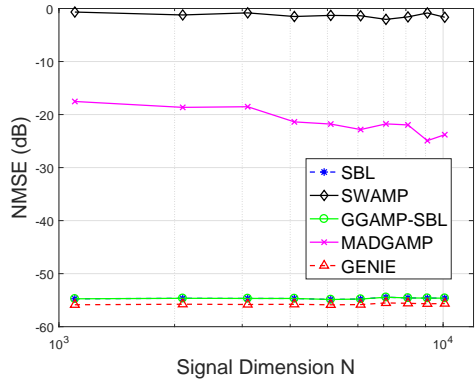}
\subcaption{NMSE versus N} \label{NMSE_Vs_N_SMV}
\end{minipage}
\begin{minipage}{\columnwidth}
\centering
\includegraphics[width=2.8in]{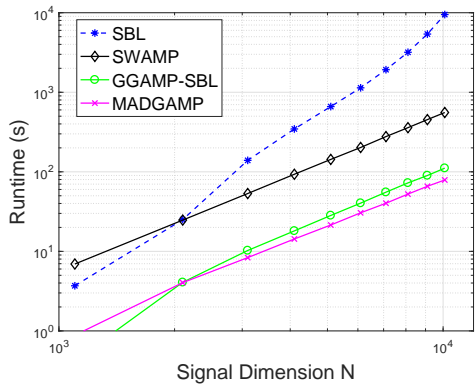}
\subcaption{Runtime versus N} \label{TIME_Vs_N_SMV}
\end{minipage}
\caption{Performance and complexity comparison for SMV algorithms versus problem dimensions}
\end{figure}

\subsubsection{Performance and complexity versus problem dimensions}
To show the effect of increasing the problem dimensions on the performance and complexity of the different algorithms, we plot the NMSE and runtime against $N$, while we keep an $M/N$ ratio of 0.5, a $K/N$ ratio of 0.2 and an SNR of 60dB. We run the experiment using column correlated matrices with $\rho=0.9$.

As expected from previous experiments, Fig.~\ref{NMSE_Vs_N_SMV} shows that only GGAMP-SBL and SBL algorithms can recover $\boldsymbol{x}$ when we use column correlated matrices with a correlation coefficient of $\rho=0.9$. The comparison between the performance of SBL and GGAMP-SBL show almost identical NMSE.\\
As problem dimensions grow, Fig.~\ref{TIME_Vs_N_SMV} shows that the difference in runtimes between the original SBL and GGAMP-SBL algorithms grows to become more significant, which suggests that the GGAMP-SBL is more practical for large size problems.

\subsubsection{Performance versus undersampling ratio $M/N$}
\begin{figure}[tbh!]
\centering
\includegraphics[width= \columnwidth]{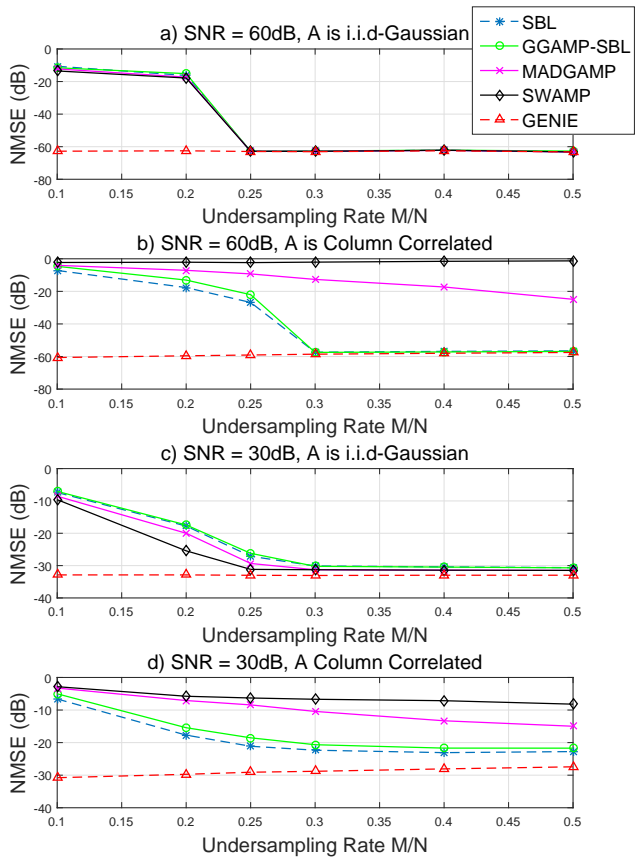}
\caption{NMSE comparison of SMV algorithms versus the undersampling rate M/N} \label{MN_plot}
\end{figure}

In this section we examine the ability of the proposed algorithm to recover a sparse vector from undersampled measurements at different undersampling ratios $M/N$.
In the below experiments we fix $N$ at 1000 and vary $M$. We set the Bernoulli-Gaussian non-zero probability $\lambda$ so that $M/K$ has an average of three measurements for each non-zero component.
We plot the NMSE versus the undersampling ratio $M/N$ for i.i.d.-Gaussian matrices $\boldsymbol{A}$ and for column correlated $\boldsymbol{A}$ with $\rho=0.9$. We run the experiments at SNR=60dB and at SNR=30dB.
In Fig.~\ref{MN_plot} we find that for SNR=60dB and i.i.d.-Gaussian $\boldsymbol{A}$, all algorithms meet the SKS bound when the undersampling ratio is larger than or equal to 0.25, while all algorithms fail to meet the bound at any ratio smaller than that. When $\boldsymbol{A}$ is column correlated, SBL and GGAMP-SBL are able to meet the SKS bound at $M/N \geq 0.3$, while MADGAMP and SwAMP do not meet the bound even at $M/N = 0.5$. We also note the MADGAMP's NMSE slowly improves with increased underasampling ratio, while SwAMP's NMSE does not.
At SNR=30dB, with i.i.d.-Gaussian $\boldsymbol{A}$ all algorithms are close to the SKS bound when the undersampling ratio is larger than 0.3. At $M/N \leq 0.3$, SBL and GGAMP-SBL are slightly outperformed by MADGAMP, while SwAMP seems to have the best performance in this region. When $\boldsymbol{A}$ is column correlated, NMSE of SBL and GGAMP-SBL outperform the other two algorithms, and similar to the SNR=60dB case, MADGAMP's NMSE seems to slowly improve with increased undersampling ratio, while SwAMP's NMSE does not improve.

\subsection{MMV GGAMP-TSBL Numerical Results}
In this section, we present  a numerical study to illustrate the performance and complexity of the proposed GGAMP-TSBL algorithm.
Although the AMP MMV algorithm in \cite{AMP_MMV} can be extended to incorporate damping, the current implementation of AMP MMV does not include damping and will diverge when used with the type of generic $\boldsymbol{A}$ matrices we are considering for our experiments. Therefore, we restrict the comparison of the performance and complexity of the GGAMP-TSBL algorithm to the TMSBL algorithm. We also compare the recovery performance against a lower bound on the achievable TNMSE by extending the support aware Kalman smoother (SKS) from \cite{AMP_MMV} to include damping and hence be able to handle generic $\boldsymbol{A}$ matrices. The implementation of the smoother is straight forward, and is exactly the same as the E-step part in Table~\ref{tab:MMV_algorithm}, when the true values of $\sigma^2$, $\boldsymbol{\gamma}$ and $\beta$ are used, and when $\boldsymbol{A}$ is modified to include only the columns corresponding to the non-zero elements in $\boldsymbol{x}^{(t)}$.
An AWGN channel was also assumed in the case of MMV.

\begin{figure}[tbh!]
\centering
\includegraphics[width=\columnwidth]{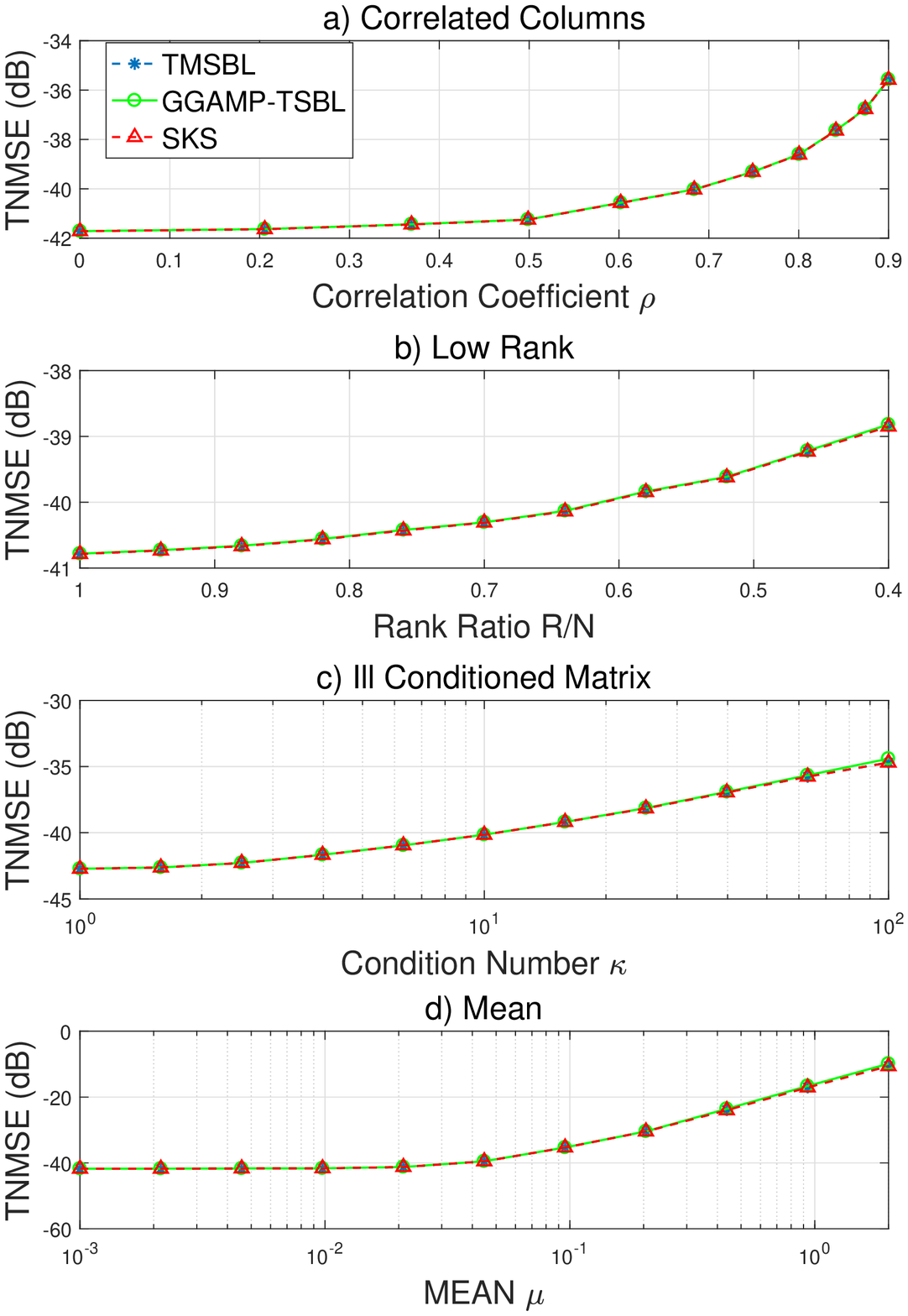}
\caption{TNMSE comparison of MMV algorithms under non-i.i.d.-Gaussian $\boldsymbol{A}$ matrices with SNR=60dB} \label{MMV_60dB_MSE}
\end{figure}
\begin{figure}[tbh!]
\centering
\includegraphics[width=\columnwidth]{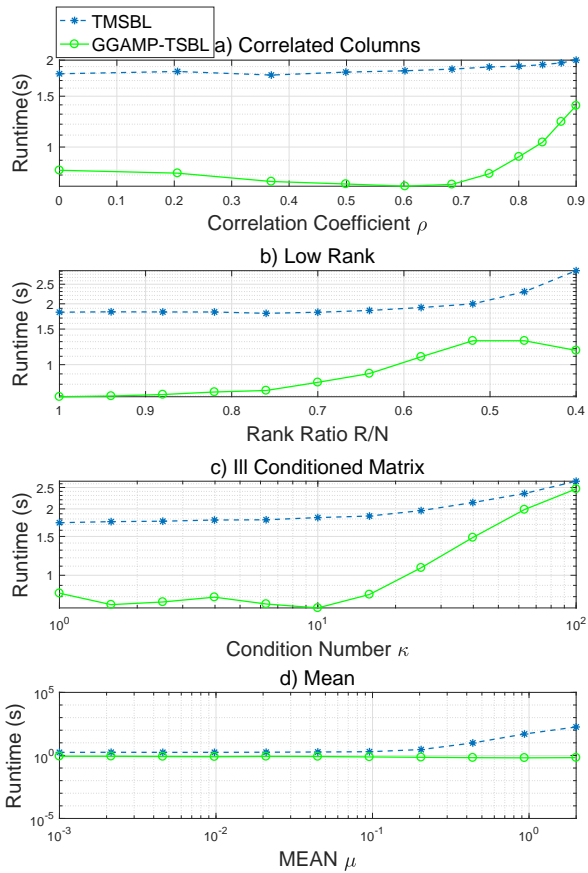}
\caption{Runtime comparison of MMV algorithms under non-i.i.d.-Gaussian $\boldsymbol{A}$ matrices with SNR=60dB} \label{MMV_60dB_Time}
\end{figure}

\subsubsection{Robustness to generic matrices at high SNR}
The experiment investigates the robustness of the proposed algorithm by comparing it to the TMSBL and the support aware smoother. Once again we use the four types of matrices mentioned at the beginning of this section, over the same range of deviation from the i.i.d.-Gaussian case. For this experiment we set $N=1000$, $M=500$, $\lambda=0.2,$ $SNR=60$dB and the temporal correlation coefficient $\beta$ to 0.9. We choose a relatively high value for $\beta$ to provide large deviation from the SMV case. This is due to the fact that the no correlation case is reduced to solving multiple SMV instances in the E-step, and then applying the M-step to update the hyperparameter vector $\boldsymbol{\gamma}$, which is common across time frames \cite{wipf2007empirical}.
The TNMSE results in Fig.~\ref{MMV_60dB_MSE} show that the performance of GGAMP-TSBL was able to match that of TMSBL in all cases and they both achieved the SKS bound.\\
Once again Fig.~\ref{MMV_60dB_Time} shows that the proposed GGAMP-TSBL was able to reduce the complexity compared to the TMSBL algorithm, even when damping was used. Although the complexity reduction does not seem to be significant for the selected problem size and SNR, we will see in the following experiments how this reduction becomes more significant as the problem size grows or as a lower SNR is used.

\subsubsection{Robustness to generic matrices at lower SNR}
\begin{figure}[tbh!]
\centering
\includegraphics[width=\columnwidth]{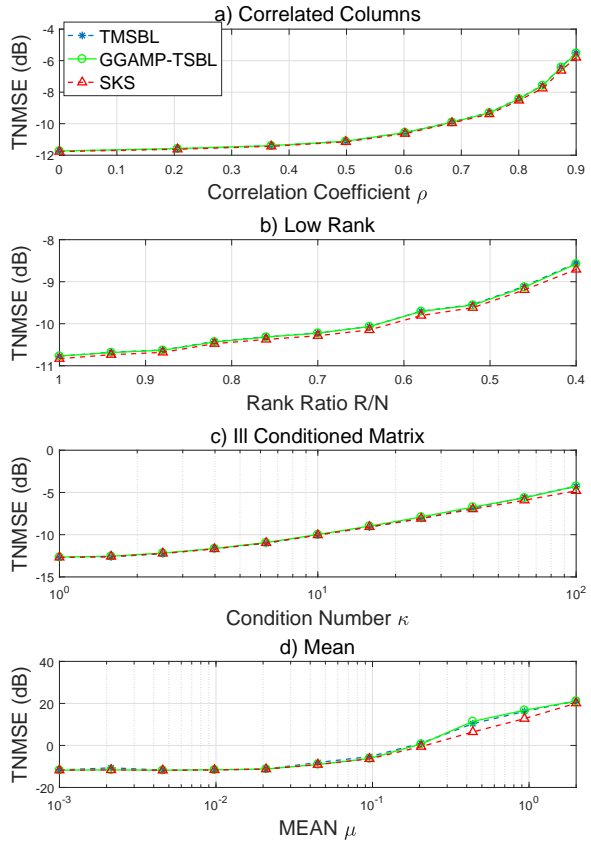}
\caption{TNSME comparison of MMV algorithms under non-i.i.d.-Gaussian $\boldsymbol{A}$ matrices with SNR=30dB} \label{MMV_30dB_MSE}
\end{figure}

\begin{figure}[tbh!]
\centering
\includegraphics[width=\columnwidth]{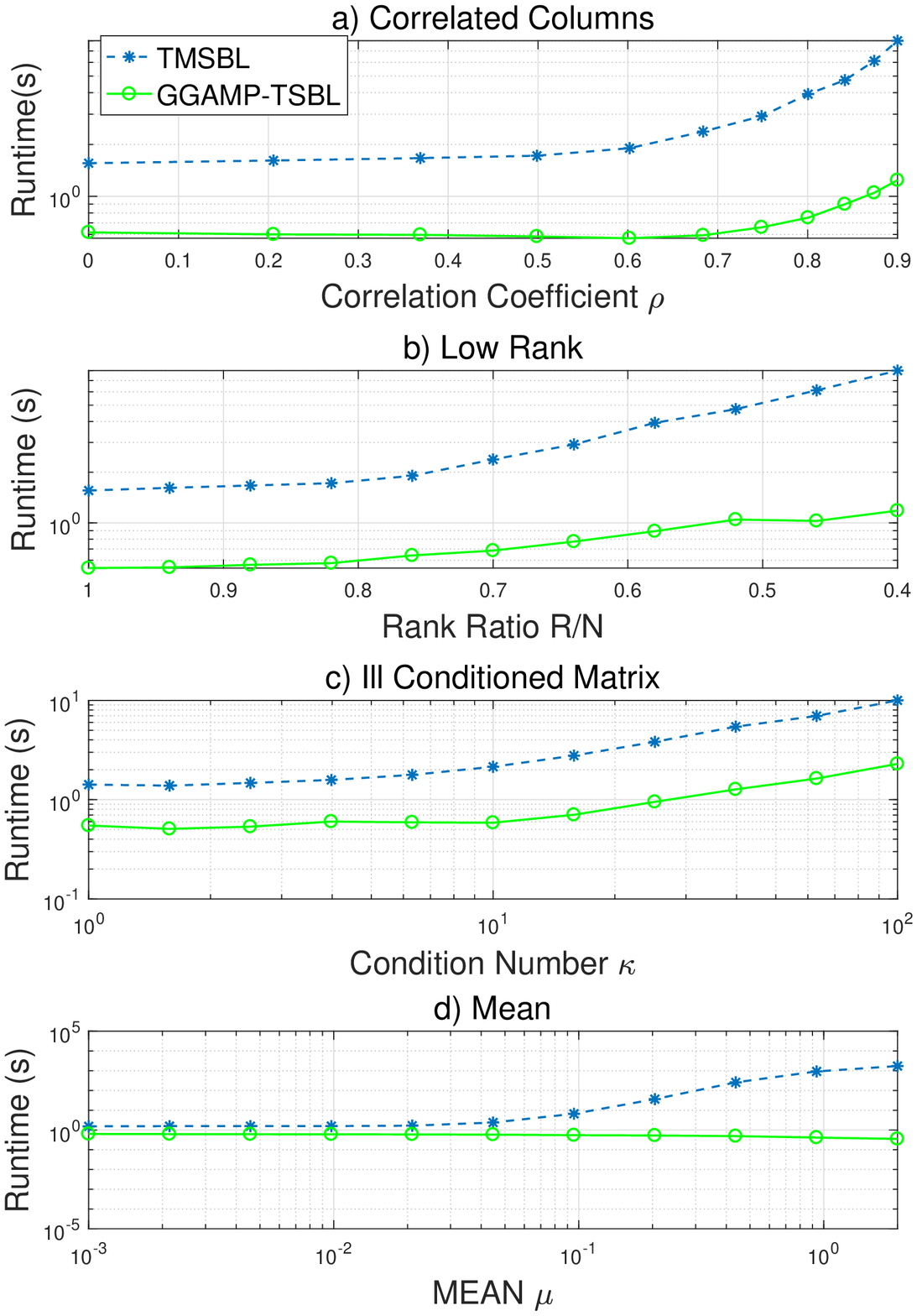}
\caption{Runtime comparison of MMV algorithms under non-i.i.d.-Gaussian $\boldsymbol{A}$ matrices with SNR=30dB} \label{MMV_30dB_Time}
\end{figure}

The performance and complexity of the proposed algorithm are examined at a lower SNR setting than the previous experiment. We set the SNR to 30dB and collect the same data points collected as in the 60dB SNR case.
Fig.~\ref{MMV_30dB_MSE} shows that the GGAMP-TSBL performance matches that of the TMSBL and almost achieves the bound in most cases.\\
Similar to the previous cases, Fig.~\ref{MMV_30dB_Time} shows that the complexity of GGAMP-TSBL is lower than that of TMSBL.

\subsubsection{Performance and complexity versus problem dimension}
\begin{figure}
\begin{minipage}{\columnwidth}
\centering
\includegraphics[width=2.8in]{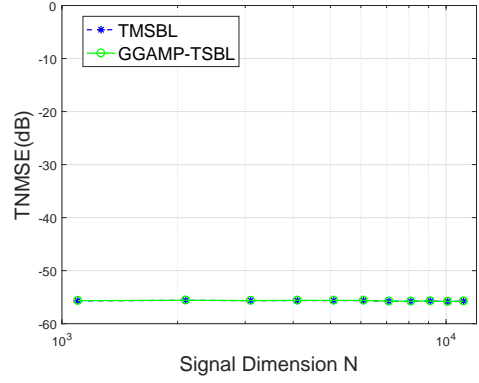}
 \subcaption{NMSE versus N} \label{TNMSE_Vs_N_MMV}
\end{minipage}
\begin{minipage}{\columnwidth}
\centering
\includegraphics[width=2.8in]{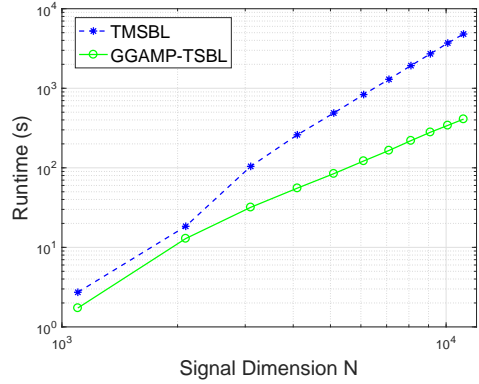}
 \subcaption{Runtime versus N} \label{Time_Vs_N_MMV}
\end{minipage}
\caption{Performance and complexity comparison for MMV algorithms versus problem dimensions}
\end{figure}

To validate the claim that the proposed algorithm is more suited to deal with large scale problems we study the algorithms' performance and complexity against the signal dimension $N$. We keep an $M/N$ ratio of 0.5, a $K/N$ ratio of 0.2 and an SNR of 60dB. We run the experiment using column correlated matrices with $\rho=0.9$. In addition, we set $\beta$ to 0.9, high temporal correlation.
In terms of performance, Fig.~\ref{TNMSE_Vs_N_MMV} shows that the proposed GGAMP-TSBL algorithm was able to match the performance of TMSBL. However, in terms of complexity, similar to the SMV case, Fig.~\ref{Time_Vs_N_MMV} shows that the runtime difference becomes more significant as the problem size grows, making the GGAMP-SBL a better choice for large scale problems.

\section{Conclusion}
In this paper, we presented a GAMP based SBL algorithm for solving the sparse signal recovery problem. SBL uses sparsity promoting priors on $\boldsymbol{x}$ that admit a Gaussian scale mixture representation.
Because of the Gaussian embedding offered by the GSM class of priors, we were able to leverage the Gaussian GAMP algorithm along with it's convergence guarantees given in \cite{Schniter_conv}, when sufficient damping is used,
to develop a reliable and fast algorithm. We numerically showed how this damped GGAMP implementation of the SBL algorithm also reduces the cost function of the original SBL approach.
The algorithm was then extended to solve the MMV SSR problem in the case of generic $\boldsymbol{A}$ matrices and temporal correlation, using a similar GAMP based SBL approach. Numerical results show that both the SMV and MMV proposed algorithms were more robust to generic $\boldsymbol{A}$ matrices when compared to other AMP algorithms. In addition, numerical results  also  show the significant reduction in complexity the proposed algorithms offer over the original SBL and TMSBL algorithms, even when sufficient damping is used to slow down the updates to guarantee convergence.
Therefore the proposed algorithms address the convergence limitations in AMP algorithms as well as the complexity challenges in traditional SBL algorithms, while retaining the positive attributes namely the robustness of SBL to generic $\boldsymbol{A}$ matrices, and the low complexity of message passing algorithms.

\appendices
\section{Derivation of GGAMP-TSBL Updates}
\subsection{The Within Step Updates}
To make the factor graph for the within step in Fig.~\ref{Message_Steps} exactly the same as the SMV factor graph we combine the product of the two messages incoming from $f_n^{(t)}$ and $f_n^{(t+1)}$ to $x_n^{(t)}$ into one message as follows:
\begin{flalign}
V_{f_n^{(t)} \rightarrow x_n^{(t)}} &\propto \mathcal{N}(x_n^{(t)}; \eta_n^{(t)}, \psi_n^{(t)}) \nonumber\\
V_{f_n^{(t+1)}\rightarrow x_n^{(t)}} &\propto \mathcal{N}(x_n^{(t)}; \theta_n^{(t)}, \phi_n^{(t)}) \nonumber\\
V_{\bar{f}_n^{(t)} \rightarrow x_n^{(t)}} &\propto \mathcal{N} (x_n^{(t)};\rho_n^{(t)},\zeta_n^{(t)}) \nonumber \\ &\propto \mathcal{N} (x_n^{(t)}; \frac{\frac{\eta_n^{(t)}}{\psi_n^{(t)}} + \frac{\theta_n^{(t)}}{\phi_n^{(t)}}}{\frac{1}{\psi_n^{(t)}}+ \frac{1}{\phi_n^{(t)}}},\frac{1}{\frac{1}{\psi_n^{(t)}}
+ \frac{1}{\phi_n^{(t)}}}). \label{eq:comb_message}
\end{flalign}
Combining these two messages reduces each time frame factor graph to an equivalent one to the SMV case with a modified prior on $x_n^{(t)}$ of \eqref{eq:comb_message}. Applying the damped GAMP algorithm from \cite{Schniter_conv} with $p(x^{(t)}_n)$ given in \eqref{eq:comb_message}:
\begin{flalign*}
g^{(t)}_x&=\frac{\frac{\boldsymbol{r}^{(t)}}{\boldsymbol{\tau}_r^{(t)}} + \frac{\boldsymbol{\rho}^{(t)}}{\boldsymbol{\zeta}^{(t)}}}{\frac{1}{\boldsymbol{\tau}_r^{(t)}} + \frac{1}{\boldsymbol{\zeta}^{(t)}}}=\frac{\frac{\boldsymbol{r}^{(t)}}{\boldsymbol{\tau}_r^{(t)}}+\frac{\boldsymbol{\eta}^{(t)}}{\boldsymbol{\psi}^{(t)}}+\frac{\boldsymbol{\theta}^{(t)}}{\boldsymbol{\phi}^{(t)}}}{\frac{1}{\boldsymbol{\tau}_r^{(t)}}+\frac{1}{\boldsymbol{\psi}^{(t)}}+\frac{1}{\boldsymbol{\phi}^{(t)}}}\\
\boldsymbol{\tau}^{(t)}_x&=\frac{1}{\frac{1}{\boldsymbol{\tau}_r^{(t)}} + \frac{1}{\boldsymbol{\zeta}^{(t)}}}=\frac{1}{\frac{1}{\boldsymbol{\tau}_r^{(t)}}+\frac{1}{\boldsymbol{\psi}^{(t)}}+\frac{1}{\boldsymbol{\phi}^{(t)}}}.
\end{flalign*}

\subsection{Forward Message Updates}
\begin{small}
\begin{flalign*}
V_{f_n^{(1)}\rightarrow x_n^{(1)}} &\propto \mathcal{N}(x_n^{(1)}; 0, \gamma_n)\\
V_{f_n^{(t)}\rightarrow x_n^{(t)}} &\propto \mathcal{N}(x_n^{(t)}; \eta_n^{(t)}, \psi_n^{(t)})\\
& \propto \int \! \left(\prod_{l=1}^M V_{g_l^{(t-1)} \rightarrow x_n^{(t-1)}} \right) V_{f_n^{(t-1)}\rightarrow x_n^{(t-1)}} \\ & \quad \quad \quad \quad  \quad \quad \quad \quad  \quad \quad \quad \quad  P(x_n^{(t)} \mathbin{\vert} x_n^{(t-1)}) \, \mathrm{d}x_n^{(t-1)}\\
& \propto \int \!  \ \mathcal{N} (x_n^{(t-1)}; r_n^{(t-1)}, \tau_{r_n}^{(t-1)}) \ \mathcal{N} (x_n^{(t-1)}; \eta_n^{(t-1)}, \psi_n^{(t-1)}) \ \\ & \quad \quad \quad \quad \quad  \quad \quad \quad \ \mathcal{N}(x_n^{(t)};\beta x_n^{(t-1)},(1-\beta^2) \gamma_n) \, \mathrm{d}x_n^{(t-1)}.
\end{flalign*}
\end{small}

Using rules for Gaussian pdf multiplication and convolution we get the $\eta_n^{(t)}$ and $\psi_n^{(t)}$ updates given in Table~\ref{tab:MMV_algorithm} equations (E3) and (E4).

\subsection{Backward Message Updates}
\begin{small}
\begin{flalign*}
V_{f_n^{(t+1)}\rightarrow x_n^{(t)}} &\propto \mathcal{N}(x_n^{(t)}; \theta_n^{(t)}, \phi_n^{(t)})\\
& \propto \int \! \left(\prod_{l=1}^M V_{g_l^{(t+1)} \rightarrow x_n^{(t+1)}} \right) V_{f_n^{(t+2)}\rightarrow x_n^{(t+1)}} \\ & \quad \quad \quad \quad  \quad \quad \quad \quad  \quad \quad \quad \quad P(x_n^{(t+1)} \mathbin{\vert} x_n^{(t)}) \, \mathrm{d}x_n^{(t+1)}\\
& \propto \int \! \ \mathcal{N} (x_n^{(t+1)}; r_n^{(t+1)}, \tau_{r_n}^{(t+1)}) \ \mathcal{N} (x_n^{(t+1)}; \theta_n^{(t+1)}, \phi_n^{(t+1)}) \ \\ & \quad \quad \quad \quad \quad  \quad \quad \quad \  \mathcal{N}(x_n^{(t+1)};\beta x_n^{(t)},(1-\beta^2) \gamma_n) \, \mathrm{d}x_n^{(t+1)}.
\end{flalign*}
\end{small}
Using rules for Gaussian pdf multiplication and convolution we get the $\theta_n^{(t)}$ and $\phi_n^{(t)}$ updates given in Table~\ref{tab:MMV_algorithm} equations (E13) and (E14).

\bibliographystyle{ieeetr}
\bibliography{c:/texmaker/work/dgamp}

\end{document}